\documentclass[10pt,conference]{IEEEtran}
\usepackage{cite}
\let\labelindent\relax
\usepackage{amsmath,amssymb,amsfonts}
\usepackage{algorithmic}
\usepackage[table,xcdraw]{xcolor}
\usepackage{textcomp}
\usepackage[framemethod=TikZ]{mdframed}
\usepackage{lipsum}
\usepackage{listings, color}
\usepackage{url}
\usepackage{hyperref}
\usepackage{float} 
\usepackage{verbatim}
\usepackage{subcaption}
\usepackage{mwe}
\usepackage{threeparttable}
\usepackage{enumitem}
\usepackage{booktabs}
\usepackage{hhline}
\usepackage{balance}
\usepackage{multirow}
\usepackage{flushend}

\usepackage[numbers,sort&compress]{natbib} 

\mdfdefinestyle{exampledefault}{
    skipabove=0.2cm,  
    middlelinecolor = black,
    middlelinewidth=0.5pt,
    linecolor=black,
    backgroundcolor=gray!5,
    roundcorner = 5pt,
}

\begin{document}

\title{Towards Understanding Fairness and its Composition in Ensemble Machine Learning\\}

\author{\IEEEauthorblockN{Usman Gohar}
\IEEEauthorblockA{\textit{Dept. of Computer Science} \\
\textit{Iowa State University}\\
Ames, IA, USA \\
ugohar@iastate.edu}
\and
\IEEEauthorblockN{Sumon Biswas}
\IEEEauthorblockA{\textit{School of Computer Science} \\
\textit{Carnegie Mellon University}\\
Pittsburgh, PA, USA \\
sumonb@cs.cmu.edu}
\and
\IEEEauthorblockN{Hridesh Rajan}
\IEEEauthorblockA{\textit{Dept. of Computer Science} \\
\textit{Iowa State University}\\
Ames, IA, USA \\
hridesh@iastate.edu}
}

\maketitle

\thispagestyle{plain}
\pagestyle{plain}

\begin{abstract}
\label{sec:abstract}
Machine Learning (ML) software has been widely adopted in modern society, with reported fairness implications for minority groups based on race, sex, age, etc. Many recent works have proposed methods to measure and mitigate algorithmic bias in ML models. The existing approaches focus on single classifier-based ML models. However, real-world ML models are often composed of multiple independent or dependent learners in an ensemble (e.g., Random Forest), where the fairness composes in a non-trivial way. How does fairness compose in ensembles? What are the fairness impacts of the learners on the ultimate fairness of the ensemble? Can fair learners result in an unfair ensemble? Furthermore, studies have shown that hyperparameters influence the fairness of ML models. Ensemble hyperparameters are more complex since they affect how learners are combined in different categories of ensembles. Understanding the impact of ensemble hyperparameters on fairness will help programmers design fair ensembles. Today, we do not understand these fully for different ensemble algorithms. In this paper, we comprehensively study popular real-world ensembles: Bagging, Boosting, Stacking, and Voting. We have developed a benchmark of 168 ensemble models collected from Kaggle on four popular fairness datasets. We use existing fairness metrics to understand the composition of fairness. Our results show that ensembles can be designed to be fairer without using mitigation techniques. We also identify the interplay between fairness composition and data characteristics to guide fair ensemble design. Finally, our benchmark can be leveraged for further research on fair ensembles. To the best of our knowledge, this is one of the first and largest studies on fairness composition in ensembles yet presented in the literature.
\end{abstract}

\begin{IEEEkeywords}
fairness, ensemble, machine learning, models
\end{IEEEkeywords}

\newcounter{finding}
\section{Introduction} 
\label{sec:introduction}

Machine learning (ML) is ubiquitous in modern software today. Due to the black-box \cite{10.1145/3338906.3338937} nature of ML algorithms and its applications in critical decision-making \cite{10.1145/3278721.3278729,mattu_machine_nodate}, fairness in ML software has become a huge concern. Measuring ML fairness \cite{10.1145/2090236.2090255, three-bayes, 10.1145/3219819.3220046, JMLR:v20:18-262} and mitigating the discrimination \cite{Chouldechova2017FairPW, three-bayes, NIPS2016_dc4c44f6}  has been studied extensively. Recent work in software engineering has shown the need to produce fair software and detect bias in complex ML software environments \cite{fairnessinsoftware, 10.1145/3287560.3287589, 10.1145/3468264.3468536, 10.1145/3368089.3409704}.

Prior research has mostly focused on fairness in standalone classifiers (e.g., \textit{Logistic Regression, SVM}) \cite{10.1145/3338906.3338937, 10.1145/3106237.3106277, article}.  However, a class of ML models called {\em ensemble models} are becoming increasingly important in practice today due to their superior performance across a multitude of ML \& real-life challenges \cite{10.5555/2627435.2697065, oza2001online, dietterich2000experimental, bauer1999empirical, HOSNI201989}, and better generalization on unseen data, especially in smaller datasets \cite{548872, 9364928,dietterich2000experimental}. Ensemble models combine the predictions of multiple base learners to make the final prediction, e.g., Random Forest uses a large number of decision trees, with the majority class being the final output. Ensemble models are the most mentioned ML algorithms on Kaggle \cite{DBLP:journals/corr/abs-2006-08334}, and in previous SE works on fairness, ensemble models comprise more than $80\%$ of the total models \cite{10.1145/3368089.3409704,10.1145/3468264.3468536}.
Like traditional ML models, ensemble models can also suffer from unfairness problem that discriminates against population subgroups based on race, gender, etc. Although many fairness mitigation techniques \cite{10.1145/3278721.3278779, 10.1145/3368089.3409697} exist, they do not always generalize well \cite{10.1145/2783258.2783311, NIPS2016_9d268236, IBMLale}. Therefore, if we better understand the fairness composition in ensembles, we can design fair ensemble models without applying mitigation techniques. In this paper, we have conducted an empirical study to understand the composition of fairness in ensembles and the interplay of their properties with fairness.

Recently, multiple works have shown that ensembles can be leveraged to enhance fairness and mitigate bias in ML models \cite{9462068, 8995403, IBMLale}. Grgic-Hlaca \MakeLowercase{\textit{et al.}} first explored fairness properties of random selection ensemble, only theoretically \cite{Nina}. Bower \MakeLowercase{\textit{et al.}} explored how fairness propagates through a multi-stage decision process like hiring \cite{article}. Similarly, Dwork \MakeLowercase{\textit{et al.}} introduced a framework to understand the composition of fairness in ensembles that \textit{only} utilize AND, OR operators to make a decision, e.g., two credit bureaus' (AND) report a score to determine loan eligibility \cite{10.1145/2090236.2090255}. Feffer \MakeLowercase{\textit{et al.}} studied how ensembles and bias mitigators can be combined using modularity to improve stability in bias mitigation \cite{IBMLale}. Therefore, it is evident that fairness in ensembles and their composition is non-trivial. Moreover, prior works in SE have shown the impact of training processes such as hyperparameter optimization, data transformation, etc., on the fairness of ML software \cite{10.1145/3510003.3510202, 10.1145/3468264.3468536, 10.1145/3368089.3409697}. We postulate that ensemble hyperparameters also impact unfairness in ensembles, and failure to study them can amplify bias. However, ensemble hyperparameters are different than typical ML model hyperparameters as they dictate the design of the ensemble, e.g., number of learners, learning method, etc. However, no empirical study has been conducted to understand fairness composition in ensembles and the effect of their hyperparameter space on fairness. To this end, we have created a benchmark of 168 real-world ensemble models from Kaggle and designed experiments to measure their fairness. We analyze fairness composition in ensemble criteria such as parallel and sequential ensembles, homogeneity of models, and different ensemble methods such as bagging, boosting, voting, stacking, etc., and all the ensemble classifiers available in the popular Scikit-learn \cite{scikit-learn}. Specifically, we answer the following overarching research questions:

\begin{itemize}
    \item \textbf{RQ1:} \textit{What are the fairness measures of various ensemble techniques?} 
    \item \textbf{RQ2:} \textit{How does fairness compose in ensemble models?}
    \item \textbf{RQ3:} \textit{Can ensemble-related hyperparameters be chosen to design fair ensemble models?}
\end{itemize}


To the best of our knowledge, this is the first work to experimentally evaluate the fairness composition in popular ensembles and elicit fair ensemble design considerations. Our results show that fairness in ensembles composes in the base learners, and fair ensemble models can be built by carefully considering the composition. The analyses also identify learners that cause fairness problems which software developers can leverage to develop frameworks to measure fairness in base learners and encourage transparency. We also identify and explore ensemble-related hyperparameters to design fair ML models for each ensemble type. Lastly, we provide a comprehensive review of fairness composition in ensembles that will help direct future research in the area. Overall, the following are the key contributions of this paper:
\begin{itemize}
    \item Explored fairness composition and its interplay with data characteristics and individual learners to mitigate bias.
    \item Empirically evaluate fairness patterns of popular ML ensemble models.
    \item We identified ensemble design considerations and hyperparameters that would guide developers in fair ensemble design and mitigate inherent unfairness effectively.
    \item A comprehensive fairness benchmark of popular ensembles that can be leveraged for further research on building fairness-aware ensembles. The benchmark, code, and experimental results are available: \url{https://github.com/UsmanGohar/FairEnsemble}
\end{itemize}

The rest of the paper is organized as follows: \S \ref{sec:motivation} describes the motivation of our work and the background on ensembles. In \S \ref{sec:methodology}, we discuss the methodology for our study, including benchmark collection, datasets and fairness and accuracy measures used, and finally, the experiment setup \& design. In \S \ref{sec:fairness-in-ensembles}, we discuss the state of fairness in ensembles \textbf{(RQ1)} and how it composes \textbf{(RQ2)}, \S \ref{sec:fair-design} discusses the design criteria to improve fairness in ensembles \textbf{(RQ3)}. Finally, we discuss the implications of our work in \S \ref{sec:discussion}, threats to validity in \S \ref{sec:threats}, related works in \S \ref{sec:related} and then present the conclusion in \S \ref{sec:conclusion}.
\section{Motivation and Background}
\label{sec:motivation}

In this section, we use a motivating example to illustrate
the complexity of fairness composition in ensembles and the need to study bias induced by certain ensemble parameters.

\subsection{Motivating example}


Ensemble models are widely deployed to win competitions in online communities like Kaggle due to their superior performances \cite{oza2001online, 10.5555/2627435.2697065, dietterich2000experimental, bauer1999empirical, HOSNI201989, DBLP:journals/corr/abs-2006-08334}. In prior SE works on fairness \cite{10.1145/3368089.3409704,10.1145/3468264.3468536}, more than 80\% of the models were ensemble based. However, those works did not consider fairness composition of individual learners, its effect on the fairness of ensembles, and the inherent bias in ensemble methods, which is non-trivial. Hence, not studying fairness composition in ensembles fails to capture the complete fairness of an ML pipeline. Consider the code snippet below of a top-performing model (Voting ensemble) from Kaggle, which is used to predict the income of an individual (\textit{German Credit} dataset). 



\definecolor{codegreen}{rgb}{0,0.6,0}
\definecolor{codegray}{rgb}{0.5,0.5,0.5}
\definecolor{codepurple}{rgb}{0.58,0,0.82}
\definecolor{codeblue}{rgb}{0, 0, 0.8}
\definecolor{backcolour}{rgb}{0.95,0.95,0.97}
\definecolor{Gray}{gray}{0.4}


\lstdefinestyle{mystyle}{
    backgroundcolor=\color{backcolour},   
    commentstyle=\color{Gray},
    keywordstyle=\color{codegreen},
    numberstyle=\tiny\color{codegray},
    stringstyle=\color{codegreen},
    basicstyle=\scriptsize,
    breakatwhitespace=false,         
    breaklines=true,                 
    captionpos=b,                    
    keepspaces=true,                 
    numbers=left,                    
    numbersep=5pt,                  
    showspaces=false,                
    showstringspaces=false,
    showtabs=false,                  
    tabsize=2,
    xleftmargin=.015\textwidth, 
    rulecolor=\color{black!15},
    otherkeywords={self, numpy.maximum, math.exp}
}
\lstset{style=mystyle}
\begin{lstlisting}[language=Python, frame=single,upquote=true]
models = []
models.append(('LGR', LogisticRegression()))
models.append(('LDA', LinearDiscriminantAnalysis()))
models.append(('KNN', KNeighborsClassifier()))
models.append(('CART', DecisionTreeClassifier()))
models.append(('NB', GaussianNB()))
models.append(('RF', RandomForestClassifier()))
models.append(('SVM', SVC(gamma='auto'))
models.append(('XGBM', XGBClassifier()))
models.append(('LGBM', LGBMClassifier()))
model = VotingClassifier(estimators=models, voting='soft')
model.fit(X_train, y_train)
y_pred = model.predict(X_test)

\end{lstlisting}

A voting ensemble is a type of {\em heterogeneous} ensemble that combines the predictions of dissimilar learners. It comprises multiple learners (lines 2-10) and uses a voting mechanism (line 11) to make the prediction. In \textit{soft} voting, the class label (1 or 0) with the higher average probability from the learners is chosen as the final prediction. We found that this ensemble is biased towards female applicants (Protected attribute: \textit{Sex}) in terms of statistical parity difference (SPD:-0.203). In this example, before training the ensemble, a developer must decide the number of learners, select which learners to use, and the voting type (\textit{soft/hard}). However, we found that ML libraries do not provide any fairness recommendations for building ensembles. Do these learners introduce unfairness in the predictions? How does the number of learners impact the fairness of the ensemble? More importantly, we observed that individual learners have their own fairness measures when analyzed in isolation but might result in an unfair model when used in an ensemble. For instance, our analysis shows that dropping XGBClassifier and LGBMClassifier (lines 9-10) can improve fairness by 27\% (SPD:-0.148). Interestingly, we discuss later how these two learners are inherently fair themselves and not responsible for the unfairness.

\begin{table*}[]
  \centering
  
  \caption{Types of ensemble models used in our experiments}
   \begin{tabular}{c|c|l|c|c}
    \hline 
    \multicolumn{1}{c|}{\textbf{Categories}} & \multicolumn{1}{c|}{\textbf{Ensemble Types}} & \multicolumn{1}{c|}{\textbf{Algorithms}} & \multicolumn{1}{c|}{\textbf{Composition}} & \multicolumn{1}{c}{\textbf{Classifiers}} \\
    \hline 
    \multirow{3}{*}{Sequential} & \multirow{3}{*}{Boosting} & \multirow{3}{*}{\shortstack[c] {Construct $n$ homogeneous estimators sequentially that 
    improve predictions \\ based on the previous estimator's 
    incorrect predictions}} & Homogeneous & XGBoost \\
          &       &       & Homogeneous & AdaBoost \\
          &       &       & Homogeneous & Gradient Boosting \\
    \hline
    \multirow{5}{*}{Parallel} & \multirow{3}{*}{Bagging} & \multirow{3}{*}{\shortstack[c] {Construct $n$ parallel homogeneous models that are aggregated using\\averaging}} & Homogeneous & Random Forest \\
         &       &       & Homogeneous & ExtraTrees \\
         &       &       & Homogeneous & Bagging Classifier \\
\cline{2-5}          & Voting & \shortstack{Construct a list of $n$ heterogeneous user-specified weighted  classifiers\\that are aggregated using majority voting or argmax} & Heterogeneous & Voting Classifier \\
\cline{2-5}          & Stacking & \shortstack[c]{Construct a list of n heterogeneous classifiers as base learners and a \\ meta-classifier to decide weights for each learner} & Heterogeneous & Stacking Classifier \\
    \hline
    \end{tabular}%
  \label{tab:ensemble-types}%
\end{table*}%

Furthermore, prior research has shown the impact of hyperparameters on fairness \cite{10.1145/3510003.3510202, 10.1145/3468264.3468536, 10.1145/3368089.3409697}. Ensemble hyperparameters dictate how ensembles combine learners for final prediction. In this example, if a developer used \textit{``hard''} voting (line 11), the fairness of the ensemble would improve (SPD: -0.195). Similarly, some of these hyperparameters also affect the design properties of the learners, which impacts fairness. \textit{XGBoost} (line 9) is another example of an ensemble (boosting). Boosting builds an ensemble of trees (learners) using various methods. What properties of these trees (e.g., tree depth, number of features, etc.) and the learning method impact the overall fairness of a boosting ensemble? Exploring these parameters will help developers understand how to design fair ensembles. Therefore, in addition to understanding fairness composition in the learners, it is equally important to understand how the design of ensembles using these parameters impacts fairness.

\subsection{Ensemble learning in ML software}
\label{subsec:ensemble-in-ml}
{\em Ensemble models} are a class of ML classifiers where the predictions from different learners (models) are pooled using a combination method (voting, average, random, etc.) to make the final predictions. In the motivational example above, we only discussed a single type of ensemble. Categories of ensembles are based on homogeneity, learning technique, and ensemble types. All the ensemble types covered in our study and the corresponding classifiers are given in Table \ref{tab:ensemble-types}. There are mainly two categories of ensembles: Sequential and Parallel.

\textbf{\textit{Sequential Ensembles.}} These ensembles sequentially generate base learners. Each learner in this ensemble depends on the previous learners in the sequence because the next learner attempts to correct the wrong predictions from the previous learner and so on \cite{8360836}. AdaBoost is an example of a sequential model where it reweighs (higher) misclassified examples.

\textbf{\textit{Parallel Ensembles.}} Parallel ensembles train individual base learners in parallel and independently of each other. These learners are combined using techniques such as bagging (a random sample of data with replacement) or voting, which encourages improved variance \cite{8360836} e.g., \textit{Random Forest.}

\textit{\textbf{Homogeneity of ensembles.}}
Ensemble methods that use single-type base learners are called \textit{homogeneous} models \cite{Elish2013EmpiricalSO}. These individual learners are combined to generate the final result, e.g., {\em XGBoost} and {\em AdaBoost} use decision trees. By contrast, \textit{heterogeneous} ensembles combine the predictions of dissimilar individual learners \cite{Elish2013EmpiricalSO}. A popular heterogeneous ensemble method is {\em Voting}. Finally, ensemble method types are divided into {\em Boosting, Bagging, Voting}, \& {\em Stacking} \cite{scikit-learn}.


\section{Methodology}
\label{sec:methodology}
In this section, we discuss the benchmark collection process, the datasets, and fairness and accuracy measures. Finally, we describe the experimental design and setup.

\begin{table}[htbp]
  \centering
  \scriptsize
  \renewcommand{\tabcolsep}{1.5pt}
  \caption{Summary of the datasets and the number of models collected for each in the benchmark }
    \begin{tabular}{c|c|r|r|r|r|r|r|r|r|r}
    \hline
    \cellcolor[HTML]{C0C0C0}\textbf{Datasets} & \cellcolor[HTML]{C0C0C0}\textbf{PA} &
    \cellcolor[HTML]{C0C0C0}\textbf{Size} &
    \cellcolor[HTML]{C0C0C0}\textbf{\#XGB} & \cellcolor[HTML]{C0C0C0}\textbf{\#ADB} & \cellcolor[HTML]{C0C0C0}\textbf{\#GBC} & \cellcolor[HTML]{C0C0C0}\textbf{\#RF} & \cellcolor[HTML]{C0C0C0}\textbf{\#ET} & \cellcolor[HTML]{C0C0C0}\textbf{\#STK} & \cellcolor[HTML]{C0C0C0}\textbf{\#VT} & \cellcolor[HTML]{C0C0C0}\textbf{Total} \\
    \hline
    \cellcolor[HTML]{EDEDED}\textbf{Adult Census} & \cellcolor[HTML]{EDEDED} sex &
    \cellcolor[HTML]{EDEDED} 32561 &
    \cellcolor[HTML]{EDEDED} 6     & \cellcolor[HTML]{EDEDED} 6     &  \cellcolor[HTML]{EDEDED} 6     & \cellcolor[HTML]{EDEDED} 6     &  \cellcolor[HTML]{EDEDED} 6     &  \cellcolor[HTML]{EDEDED} 1     & \cellcolor[HTML]{EDEDED} 2     & \cellcolor[HTML]{EDEDED} 33 \\
    \textbf{Titanic ML} & sex  
    & 891 & 6     & 6     & 6     & 6     & 6     & 6     & 6     & 42 \\
    \cellcolor[HTML]{EDEDED}\textbf{Bank Marketing} & \cellcolor[HTML]{EDEDED} age   &
    \cellcolor[HTML]{EDEDED} 41118 &
    \cellcolor[HTML]{EDEDED} 6     & \cellcolor[HTML]{EDEDED} 6     & \cellcolor[HTML]{EDEDED} 6     & \cellcolor[HTML]{EDEDED} 6     & \cellcolor[HTML]{EDEDED} 2     &  \cellcolor[HTML]{EDEDED} 1     & \cellcolor[HTML]{EDEDED} 2     & \cellcolor[HTML]{EDEDED} 29 \\
    \textbf{German Credit} & sex   & 1000 & 6     & 1     & 1     & 6     & 1     & 1     & 1     & 17 \\
    \hline
    \end{tabular}%
  \label{tab:datasetsummary}%
  \begin{tablenotes}
      \centering
      \item  XGB: XGBoost, ADB: AdaBoost, GBC: Gradient Boosting
      , RF: Random Forest, ET: Extra Trees, STK: Stacking, VT: Voting
    \end{tablenotes}
\end{table}%

\subsection{Benchmark Collection}
\label{subsec:benchmark-collection}

For our experiments, we collected ensemble models from Kaggle \cite{kaggle} for datasets that have been used in prior fairness literature \cite{10.1145/3468264.3468537, 10.1145/3468264.3468536, IBMLale}. Unlike these works, we only collect ensemble-based models for evaluation. Specifically, we collect all ensemble classifiers available via the popular scikit-learn library \cite{scikit-learn}. We follow a similar benchmark collection process as in \cite{10.1145/3368089.3409704}. Table \ref{tab:datasetsummary} summarizes the datasets and the classifiers in the benchmark. 

For each dataset, we collected Kaggle kernels for each ensemble type in Table \ref{tab:ensemble-types} and classifiers available in scikit-learn. We filter these kernels based on four-step filtering criteria: $1)$ it should contain the predictive model (some kernels focus on data exploration only), $2)$ protected attribute is included in the training data, $3)$ at least five up-votes, and $4)$ ranked by up-votes. We used Kaggle API to collect these models and pass them through the filtering criteria. Finally, we select the top 6 models for each ensemble classifier from each dataset. In total, we created a benchmark of 168 ensembles across four datasets. We could not find certain classifiers on Kaggle for datasets like German Credit. To handle those, we use default models from scikit-learn to ensure we can evaluate across different datasets. The number of models mined is shown in Table \ref{tab:datasetsummary}. Next, we present an overview of the datasets used in our benchmark.

\textit{Adult Census}. The dataset contains income and personal information about individuals \cite{adult-kaggle}. We used \textit{sex} as the protected attribute and \textit{male} as the privileged class. The classification task predicts if a person makes over \$50,000 in annual income.

\textit{Titanic}. The dataset contains passenger data, such as gender, cabin class, etc., and is pre-split into train \& test sets; however, the test set does not contain any instance of a male passenger surviving \cite{titanic-kaggle}. Hence, we only use the training set, with \textit{gender} as the protected attribute and \textit{Female} as the privileged class. The prediction task is whether a passenger survives.

\textit{Bank Marketing}. The dataset contains bank customers' personal information such as age, job type, etc. \cite{bank-marketing-kaggle}. The protected attribute is \textit{age}, where \textit{age $> 25$} is considered privileged class and \textit{age $< 25$} as unprivileged \cite{10.1145/3368089.3409704}. The prediction task determines whether a client will subscribe to a term deposit.

\textit{German Credit}. This dataset contains personal and financial information about individuals who apply for loans at a bank \cite{german-credit}. We used the processed dataset \cite{german-credi-kaggle} since most models in our benchmark used it. This version has nine attributes, such as sex, credit amount, etc. We choose \textit{sex} as the protected attribute and \textit{male} as the privileged class. The prediction task is whether an individual is a credit risk.

\subsection{Measures}
\label{subsec:measures}

Multiple quantitative fairness and accuracy measures are available to evaluate a model. We use measures that have been previously used in literature \cite{10.1145/3287560.3287589, 10.1145/3368089.3409704}. Let $D = (T,S,Y)$ be a dataset where $T$ is the training set, $S$ is the protected attribute ($S = 1$, if privileged group (\textit{p}), else $S = 0$ (\textit{up})) and $Y$ is the classification label ($Y = 1$ if favorable label, else $Y = 0$). Let $\hat{Y}$ denote the prediction of an ML model. Next, we define our measures in terms of these notations.

\subsubsection{Accuracy Metrics}
\label{subsubsec:accuracy-measures}

We evaluate the performance of the models using accuracy and F1 metrics as defined below:
\begin{align*}
    &Accuracy = (true\ positive + true\ negative) / total \\
    &F1 = 2 *(precision * recall)/(precision + recall)
\end{align*}

where recall: $TP/(TP + FN)$, precision: $TP/(TP+FP)$ \\

\subsubsection{Fairness Measures}
\label{subsubsec:fairness-measures}

Broadly, fairness metrics are divided into three categories \cite{Binns2018FairnessIM}. We have selected a subset of these metrics representing the three categories without being exhaustive. Furthermore, we have followed the recommendations of Friedler \textit{et al.} \cite{10.1145/3368089.3409704} in terms of metrics selection. \\

\noindent \textbf{Group fairness metrics:}
\textit{Group fairness} means similar predictive outcomes for protected attributes, e.g., race (Asian/White) on a group level.

\noindent \textit{Equal Opportunity Difference (EOD):} This is defined as the difference of true-positive rates (TPR) between privileged and unprivileged groups \cite{Bellamy2018AIF3}.
\begin{align*}
    EOD = TPR_{up} - TPR_p
\end{align*}
where TPR: $TP/(TP + FN)$, FPR:  $FP/(FP + TN)$ \\

\noindent \textit{Average Odds Difference (AOD):} This is defined as the mean of false-positive rate (FPR) difference and true-positive rate difference between unprivileged and privileged groups \cite{NIPS2016_9d268236}.

\begin{center}
    $AOD = {[(FPR_{up} - FPR_p) + (TPR_{up} - TPR_p)}]/2$ 
\end{center} 

\noindent \textit{Disparate Impact (DI):} This is defined as the ratio of the probability of unprivileged group vs. privileged group getting a favorable prediction \cite{JMLR:v20:18-262}

\begin{center}
    $DI = P[\hat{S} = 1| Y = 0]/ P[\hat{S} = 1| Y = 1]$
\end{center} 

\noindent We convert Disparate Impact (DI) to log scale to improve readability compared with other metrics. \\

\noindent \textit{Statistical Parity Difference (SPD):} This is defined similar to DI but uses the difference between the probabilities. \cite{Calders2010ThreeNB}.

\begin{center}
    $SPD = P[\hat{S} = 1| Y = 0] - P[\hat{S} = 1| Y = 1]$
\end{center}

\noindent \textbf{Individual fairness metrics:}

\noindent \textit{Theil Index (TI):} It measures both the group and individual fairness \cite{10.1145/3219819.3220046}. It is defined using the following equation: 
\begin{center}

$TI = \sum_{i=1}^{n} \frac{b_i}{\Bar{a}} ln \frac{b_i}{\Bar{a}}$, where $b_i = \hat{s}_i - s_i + 1$.
\end{center}

\subsection{Experiment Design \& Setup}
\label{subsec:experiment-design-setup}

Each ensemble model has specific requirements for training (e.g., \textit{XGBoost} can handle Null values, but \textit{Random Forest} cannot \textit{etc.}) that we need to handle before we can evaluate them. We used the same preprocessing steps across all the kernels and datasets to ensure consistent comparison. Next, we evaluated the accuracy and fairness of base learners and the final ensemble level and analyzed the results.

For our data preprocessing, we start by converting all non-numerical features to categorical data, i.e., \textit{Binary} or \textit{Ordinal} (e.g., $\textit{{male}: 1}, \textit{{female}: 0}$ or  non-binary levels, like \textit{Marital-Status} to \textit{{Divorced: 0}, {Married: 1}, {Single: 2}} etc.). Next, we remove missing values from our datasets and convert continuous sensitive attributes to categorical (e.g., age $>25$: 1, age $<25$ :0 corresponding to old and young, respectively). These preprocessing steps are necessary for most ensembles and the AIF360 toolkit. We denote the privileged and unprivileged groups and the favorable label for each dataset separately. For example, in \textit{Titanic} dataset, \textit{male} is the unprivileged group, and the favorable label is {\textit{Survivied: 1}} i.e., the individual survived the titanic crash. The groups and the labels have been chosen as seen before in literature \cite{martinez2019fairness, 10.1145/3368089.3409704}. Finally, the dataset is shuffled and split into train and test sets using a $70\%-30\%$ split.
For each dataset, we have selected the top 6 kernels by upvotes. We run the preprocessing steps discussed before training the model to evaluate based on accuracy and fairness metrics. We use five fairness metrics and two accuracy measures to generate results for each model. These experiments are repeated ten times, and the mean is reported \cite{10.1145/3287560.3287589}. We used the \textit{IBM AIF 360 Fairness Toolkit} to calculate the fairness metrics.
Finally, a non-zero value for fairness metrics suggests a bias in the model. A positive value of a fairness metric suggests the model is biased against the privileged group and vice-versa.

\definecolor{darkestblue}{HTML}{329a9d}
\definecolor{darkerblue}{HTML}{68cbd0}
\definecolor{lightblue}{HTML}{96fffb}
\definecolor{lighterblue}{HTML}{d3fefd}

\begin{table*}[]
\centering
\caption{Fairness and accuracy comparison of all ensemble ML classifiers across the datasets in our benchmark. The ranks
were calculated using the Scott-Knott test \cite{7194626}. Each cell depicts the median score;\colorbox{darkestblue}{Darker}, \colorbox{darkerblue}{lighter}, \colorbox{lightblue}{light}, \colorbox{lighterblue}{lightest} and white
colored cell denotes the first, the second, the third, fourth, and lowest rank, respectively. The rank ranges from 1 to 5.}
\setlength{\tabcolsep}{2.6pt} 
\setlength\extrarowheight{1pt}
\begin{tabular}{|c|c|c|c|r|r|r|r|r|r|r|c|c|}
\hline
\rowcolor[HTML]{DAE8FC} 
Dataset                                                 & \begin{tabular}[c]{@{}c@{}}Protected\\ Attribute\end{tabular} & \begin{tabular}[c]{@{}c@{}}Ensemble\\ Classifiers\end{tabular} & \begin{tabular}[c]{@{}c@{}}Ensemble\\ Type\end{tabular} & Accuracy (+)                 & F1 (+)                       & SPD (-)                       & EOD (-)                       & AOD (-)                      & DI (-)                        & TI (-)                       & \begin{tabular}[c]{@{}c@{}}Mean Accuracy\\ Rank (-)\end{tabular} & \begin{tabular}[c]{@{}c@{}}Mean Fairness\\ Rank (-)\end{tabular} \\ \hline
                                                        & \cellcolor[HTML]{FFFFFF}                                      & \cellcolor[HTML]{FFFFFF}TM-XGB                                & \cellcolor[HTML]{FFFFFF}                                & \cellcolor[HTML]{68CBD0}0.82 & \cellcolor[HTML]{68CBD0}0.75 & \cellcolor[HTML]{329A9D}-0.65 & \cellcolor[HTML]{329A9D}-0.50 & \cellcolor[HTML]{329A9D}0.43 & \cellcolor[HTML]{329A9D}-1.83 & \cellcolor[HTML]{96FFFB}0.14 & 2                                                                & 1.4                                                              \\ \cline{3-3} \cline{5-13} 
                                                        & \cellcolor[HTML]{FFFFFF}                                      & \cellcolor[HTML]{FFFFFF}TM-ADB                                & \cellcolor[HTML]{FFFFFF}                                & 0.81                         & \cellcolor[HTML]{96FFFB}0.75 & -0.81                         & -0.77                         & 0.70                         & \cellcolor[HTML]{FFFFFF}-2.56 & \cellcolor[HTML]{68CBD0}0.15 & 3.5                                                              & 4.4                                                              \\ \cline{3-3} \cline{5-13} 
                                                        & \cellcolor[HTML]{FFFFFF}                                      & \cellcolor[HTML]{FFFFFF}TM-GBC                                & \multirow{-3}{*}{\cellcolor[HTML]{FFFFFF}Boosting}      & \cellcolor[HTML]{96FFFB}0.82 & \cellcolor[HTML]{D3FEFD}0.74 & \cellcolor[HTML]{68CBD0}-0.71 & \cellcolor[HTML]{96FFFB}-0.57 & \cellcolor[HTML]{68CBD0}0.54 & \cellcolor[HTML]{68CBD0}-2.09 & \cellcolor[HTML]{68CBD0}0.14 & 3.5                                                              & 2.2                                                              \\ \cline{3-13} 
                                                        & \cellcolor[HTML]{FFFFFF}                                      & \cellcolor[HTML]{FFFFFF}TM-RF                                 & \cellcolor[HTML]{FFFFFF}                                & \cellcolor[HTML]{D3FEFD}0.81 & 0.73                         & \cellcolor[HTML]{96FFFB}-0.68 & \cellcolor[HTML]{D3FEFD}-0.58 & \cellcolor[HTML]{96FFFB}0.52 & \cellcolor[HTML]{96FFFB}-2.30 & \cellcolor[HTML]{329A9D}0.16 & 5                                                                & 3                                                                \\ \cline{3-3} \cline{5-13} 
                                                        & \cellcolor[HTML]{FFFFFF}                                      & \cellcolor[HTML]{FFFFFF}TM-ET                                 & \multirow{-2}{*}{\cellcolor[HTML]{FFFFFF}Bagging}       & \cellcolor[HTML]{68CBD0}0.82 & \cellcolor[HTML]{68CBD0}0.75 & \cellcolor[HTML]{D3FEFD}-0.80 & -0.75                         & \cellcolor[HTML]{D3FEFD}0.68 & \cellcolor[HTML]{D3FEFD}-2.76 & \cellcolor[HTML]{96FFFB}0.15 & 2                                                                & 4                                                                \\ \cline{3-13} 
                                                        & \cellcolor[HTML]{FFFFFF}                                      & TM-VT                                                         & Voting                                                  & \cellcolor[HTML]{329A9D}0.83 & \cellcolor[HTML]{329A9D}0.77 & \cellcolor[HTML]{96FFFB}-0.74 & \cellcolor[HTML]{68CBD0}-0.51 & \cellcolor[HTML]{68CBD0}0.54 & \cellcolor[HTML]{68CBD0}-2.10 & 0.12                         & 1                                                                & 2.8                                                              \\ \cline{3-13} 
\multirow{-7}{*}{Titanic}                               & \multirow{-7}{*}{\cellcolor[HTML]{FFFFFF}Sex}                 & TM-STK                                                        & Stacking                                                & \cellcolor[HTML]{68CBD0}0.82 & \cellcolor[HTML]{96FFFB}0.76 & \cellcolor[HTML]{96FFFB}-0.76 & \cellcolor[HTML]{96FFFB}-0.63 & \cellcolor[HTML]{96FFFB}0.58 & \cellcolor[HTML]{96FFFB}-2.40 & \cellcolor[HTML]{329A9D}0.13 & 3                                                                & 2.6                                                              \\ \hline
                                                        & \cellcolor[HTML]{FFFFFF}                                      & \cellcolor[HTML]{FFFFFF}AC-XGB                                & \cellcolor[HTML]{FFFFFF}                                & \cellcolor[HTML]{329A9D}0.87 & \cellcolor[HTML]{329A9D}0.71 & \cellcolor[HTML]{D3FEFD}-0.18 & \cellcolor[HTML]{329A9D}-0.08 & \cellcolor[HTML]{329A9D}0.08 & \cellcolor[HTML]{68CBD0}-1.14 & \cellcolor[HTML]{329A9D}0.11 & 1                                                                & 1.6                                                              \\ \cline{3-3} \cline{5-13} 
                                                        & \cellcolor[HTML]{FFFFFF}                                      & \cellcolor[HTML]{FFFFFF}AC-ADB                                & \cellcolor[HTML]{FFFFFF}                                & \cellcolor[HTML]{96FFFB}0.86 & \cellcolor[HTML]{68CBD0}0.66 & \cellcolor[HTML]{FFFFFF}-0.20 & \cellcolor[HTML]{FFFFFF}-0.15 & \cellcolor[HTML]{FFFFFF}0.14 & -1.32                         & \cellcolor[HTML]{68CBD0}0.12 & 2.5                                                              & 4.4                                                              \\ \cline{3-3} \cline{5-13} 
                                                        & \cellcolor[HTML]{FFFFFF}                                      & \cellcolor[HTML]{FFFFFF}AC-GBC                                & \multirow{-3}{*}{\cellcolor[HTML]{FFFFFF}Boosting}      & \cellcolor[HTML]{68CBD0}0.86 & \cellcolor[HTML]{68CBD0}0.68 & \cellcolor[HTML]{D3FEFD}-0.19 & \cellcolor[HTML]{D3FEFD}-0.14 & \cellcolor[HTML]{68CBD0}0.11 & \cellcolor[HTML]{96FFFB}-1.25 & \cellcolor[HTML]{68CBD0}0.12 & 2                                                                & 3                                                                \\ \cline{3-13} 
                                                        & \cellcolor[HTML]{FFFFFF}                                      & \cellcolor[HTML]{FFFFFF}AC-RF                                 & \cellcolor[HTML]{FFFFFF}                                & 0.85                         & 0.67                         & \cellcolor[HTML]{68CBD0}-0.18 & \cellcolor[HTML]{D3FEFD}-0.13 & \cellcolor[HTML]{96FFFB}0.11 & \cellcolor[HTML]{D3FEFD}-1.26 & \cellcolor[HTML]{D3FEFD}0.12 & 5                                                                & 3.4                                                              \\ \cline{3-3} \cline{5-13} 
                                                        & \cellcolor[HTML]{FFFFFF}                                      & \cellcolor[HTML]{FFFFFF}AC-ET                                 & \multirow{-2}{*}{\cellcolor[HTML]{FFFFFF}Bagging}       & \cellcolor[HTML]{FFFFFF}0.84 & \cellcolor[HTML]{96FFFB}0.65 & \cellcolor[HTML]{96FFFB}-0.19 & \cellcolor[HTML]{68CBD0}-0.10 & \cellcolor[HTML]{D3FEFD}0.10 & \cellcolor[HTML]{329A9D}-1.11 & \cellcolor[HTML]{D3FEFD}0.13 & 4                                                                & 2.8                                                              \\ \cline{3-13} 
                                                        & \cellcolor[HTML]{FFFFFF}                                      & AC-VT                                                         & Voting                                                  & \cellcolor[HTML]{D3FEFD}0.85 & 0.66                         & \cellcolor[HTML]{329A9D}-0.17 & \cellcolor[HTML]{D3FEFD}-0.13 & \cellcolor[HTML]{96FFFB}0.09 & \cellcolor[HTML]{D3FEFD}-1.29 & 0.12                         & 4.5                                                              & 3.4                                                              \\ \cline{3-13} 
\multirow{-7}{*}{Adult}                                 & \multirow{-7}{*}{\cellcolor[HTML]{FFFFFF}Sex}                 & AC-STK                                                        & Stacking                                                & \cellcolor[HTML]{68CBD0}0.86 & \cellcolor[HTML]{96FFFB}0.68 & \cellcolor[HTML]{68CBD0}-0.18 & \cellcolor[HTML]{96FFFB}-0.11 & \cellcolor[HTML]{D3FEFD}0.09 & \cellcolor[HTML]{D3FEFD}-1.28 & \cellcolor[HTML]{68CBD0}0.11 & 2.5                                                              & 3.6                                                              \\ \hline
                                                        & \cellcolor[HTML]{FFFFFF}                                      & \cellcolor[HTML]{FFFFFF}BM-XGB                                & \cellcolor[HTML]{FFFFFF}                                & \cellcolor[HTML]{68CBD0}0.93 & \cellcolor[HTML]{68CBD0}0.70 & \cellcolor[HTML]{68CBD0}0.15  & \cellcolor[HTML]{96FFFB}0.08  & \cellcolor[HTML]{96FFFB}0.08 & \cellcolor[HTML]{68CBD0}0.77  & \cellcolor[HTML]{329A9D}0.05 & 1.5                                                              & 2.2                                                              \\ \cline{3-3} \cline{5-13} 
                                                        & \cellcolor[HTML]{FFFFFF}                                      & \cellcolor[HTML]{FFFFFF}BM-ADB                                & \cellcolor[HTML]{FFFFFF}                                & \cellcolor[HTML]{D3FEFD}0.88 & 0.49                         & \cellcolor[HTML]{68CBD0}0.15  & \cellcolor[HTML]{FFFFFF}0.18  & \cellcolor[HTML]{FFFFFF}0.12 & \cellcolor[HTML]{FFFFFF}1.04  & 0.11                         & 4.5                                                              & 4.4                                                              \\ \cline{3-3} \cline{5-13} 
                                                        & \cellcolor[HTML]{FFFFFF}                                      & \cellcolor[HTML]{FFFFFF}BM-GBC                                & \multirow{-3}{*}{\cellcolor[HTML]{FFFFFF}Boosting}      & \cellcolor[HTML]{D3FEFD}0.89 & 0.48                         & \cellcolor[HTML]{68CBD0}0.14  & \cellcolor[HTML]{D3FEFD}0.12  & \cellcolor[HTML]{96FFFB}0.09 & 1.09                          & 0.10                         & 4.5                                                              & 3.8                                                              \\ \cline{3-13} 
                                                        & \cellcolor[HTML]{FFFFFF}                                      & \cellcolor[HTML]{FFFFFF}BM-RF                                 & \cellcolor[HTML]{FFFFFF}                                & \cellcolor[HTML]{D3FEFD}0.89 & \cellcolor[HTML]{96FFFB}0.55 & 0.18                          & \cellcolor[HTML]{96FFFB}0.09  & \cellcolor[HTML]{96FFFB}0.09 & \cellcolor[HTML]{68CBD0}0.80  & \cellcolor[HTML]{68CBD0}0.07 & 2.5                                                              & 3                                                                \\ \cline{3-3} \cline{5-13} 
                                                        & \cellcolor[HTML]{FFFFFF}                                      & \cellcolor[HTML]{FFFFFF}BM-ET                                 & \multirow{-2}{*}{\cellcolor[HTML]{FFFFFF}Bagging}       & \cellcolor[HTML]{96FFFB}0.91 & \cellcolor[HTML]{96FFFB}0.54 & \cellcolor[HTML]{68CBD0}0.14  & \cellcolor[HTML]{68CBD0}0.06  & \cellcolor[HTML]{68CBD0}0.06 & \cellcolor[HTML]{68CBD0}0.82  & \cellcolor[HTML]{96FFFB}0.07 & 2.5                                                              & 2.2                                                              \\ \cline{3-13} 
                                                        & \cellcolor[HTML]{FFFFFF}                                      & BM-VT                                                         & Voting                                                  & \cellcolor[HTML]{329A9D}0.94 & \cellcolor[HTML]{68CBD0}0.69 & \cellcolor[HTML]{329A9D}0.12  & \cellcolor[HTML]{68CBD0}0.06  & \cellcolor[HTML]{329A9D}0.05 & \cellcolor[HTML]{329A9D}0.71  & \cellcolor[HTML]{329A9D}0.05 & 1                                                                & 1.2                                                              \\ \cline{3-13} 
\multirow{-7}{*}{Bank Marketing}                        & \multirow{-7}{*}{\cellcolor[HTML]{FFFFFF}Age}                 & BM-STK                                                        & Stacking                                                & \cellcolor[HTML]{329A9D}0.93 & \cellcolor[HTML]{329A9D}0.72 & \cellcolor[HTML]{68CBD0}0.15  & \cellcolor[HTML]{329A9D}0.04  & \cellcolor[HTML]{329A9D}0.06 & \cellcolor[HTML]{329A9D}0.71  & \cellcolor[HTML]{329A9D}0.05 & 1.5                                                              & 1.2                                                              \\ \hline
\cellcolor[HTML]{FFFFFF}                                & \cellcolor[HTML]{FFFFFF}                                      & \cellcolor[HTML]{FFFFFF}GC-XGB                                &                                                         & \cellcolor[HTML]{68CBD0}0.72 & \cellcolor[HTML]{329A9D}0.65
 & \cellcolor[HTML]{329A9D}-0.07
 & \cellcolor[HTML]{329A9D}-0.02
 & \cellcolor[HTML]{329A9D}0.08
 & \cellcolor[HTML]{329A9D}-0.12

 & 0.17
                         & 1.5                                                                & 1.8                                                             \\ \cline{3-3} \cline{5-13} 
\cellcolor[HTML]{FFFFFF}                                & \cellcolor[HTML]{FFFFFF}                                      & \cellcolor[HTML]{FFFFFF}GC-ADB                                &                                                         & \cellcolor[HTML]{96FFFB}0.72
 & \cellcolor[HTML]{96FFFB}0.55

                        & -0.11
                         & -0.07
 & \cellcolor[HTML]{D3FEFD}0.13
                         & -0.19

 & \cellcolor[HTML]{96FFFB}0.16
 & 3                                                                & 4.4                                                               \\ \cline{3-3} \cline{5-13} 
\cellcolor[HTML]{FFFFFF}                                & \cellcolor[HTML]{FFFFFF}                                      & \cellcolor[HTML]{FFFFFF}GC-GBC                                & \multirow{-3}{*}{Boosting}                              & \cellcolor[HTML]{96FFFB}0.72
 & \cellcolor[HTML]{96FFFB}0.56
                         & \cellcolor[HTML]{68CBD0}-0.08
 & \cellcolor[HTML]{D3FEFD}-0.06
 & \cellcolor[HTML]{68CBD0}0.10
 & \cellcolor[HTML]{96FFFB}-0.15
 & \cellcolor[HTML]{68CBD0}0.15
 & 3                                                              & 2.6                                                             \\ \cline{3-13} 
\cellcolor[HTML]{FFFFFF}                                & \cellcolor[HTML]{FFFFFF}                                      & \cellcolor[HTML]{FFFFFF}GC-RF                                 &                                                         & \cellcolor[HTML]{96FFFB}0.72
 & \cellcolor[HTML]{329A9D}0.64

 & \cellcolor[HTML]{68CBD0}-0.09
 & \cellcolor[HTML]{68CBD0}-0.04
 & \cellcolor[HTML]{68CBD0}0.09
 & \cellcolor[HTML]{68CBD0}-0.13
 & \cellcolor[HTML]{68CBD0}0.15

 & 2                                                             & 2                                                                \\ \cline{3-3} \cline{5-13} 
\cellcolor[HTML]{FFFFFF}                                & \cellcolor[HTML]{FFFFFF}                                      & \cellcolor[HTML]{FFFFFF}GC-ET                                 & \multirow{-2}{*}{Bagging}                               & \cellcolor[HTML]{D3FEFD}0.70
 & \cellcolor[HTML]{68CBD0}0.60

 & \cellcolor[HTML]{D3FEFD}-0.11
                         & -0.07
                         & \cellcolor[HTML]{96FFFB}0.11
                         & \cellcolor[HTML]{68CBD0}-0.14
                         & \cellcolor[HTML]{96FFFB}0.16
 & 3                                                              & 3.4                                                              \\ \cline{3-13} 
\cellcolor[HTML]{FFFFFF}                                & \cellcolor[HTML]{FFFFFF}                                      & GC-VT                                                         & Voting                                                  & \cellcolor[HTML]{329A9D}0.73
 & \cellcolor[HTML]{96FFFB}0.54

                         & \cellcolor[HTML]{68CBD0}-0.08

 & \cellcolor[HTML]{96FFFB}-0.05

 & \cellcolor[HTML]{329A9D}0.08

 & \cellcolor[HTML]{D3FEFD}-0.16

 & \cellcolor[HTML]{96FFFB}0.16

 & 2                                                                & 2.6                                                              \\ \cline{3-13} 
\multirow{-7}{*}{\cellcolor[HTML]{FFFFFF}German Credit} & \multirow{-7}{*}{\cellcolor[HTML]{FFFFFF}Sex}                 & GC-STK                                                        & Stacking                                                & \cellcolor[HTML]{329A9D}0.73
                         & \cellcolor[HTML]{96FFFB}0.55
                        & \cellcolor[HTML]{96FFFB}-0.09
 & \cellcolor[HTML]{D3FEFD}-0.06
& \cellcolor[HTML]{68CBD0}0.10
& \cellcolor[HTML]{D3FEFD}-0.17 
& \cellcolor[HTML]{329A9D} 0.14
                         & 2                                                                & 2.8                                                                \\ \hline
\end{tabular}
\label{Table_Comparison_1}
\end{table*}

\section{Fairness in Ensembles and its composition}
\label{sec:fairness-in-ensembles}

In this section, we explore the state of fairness in ensembles and its composition in all popular ensemble methods.

\subsection{State of fairness in ensemble models}
\label{subsubsec:state-of-fairness}

\begin{figure*}[t]
  \centering
    \includegraphics[width=\linewidth]{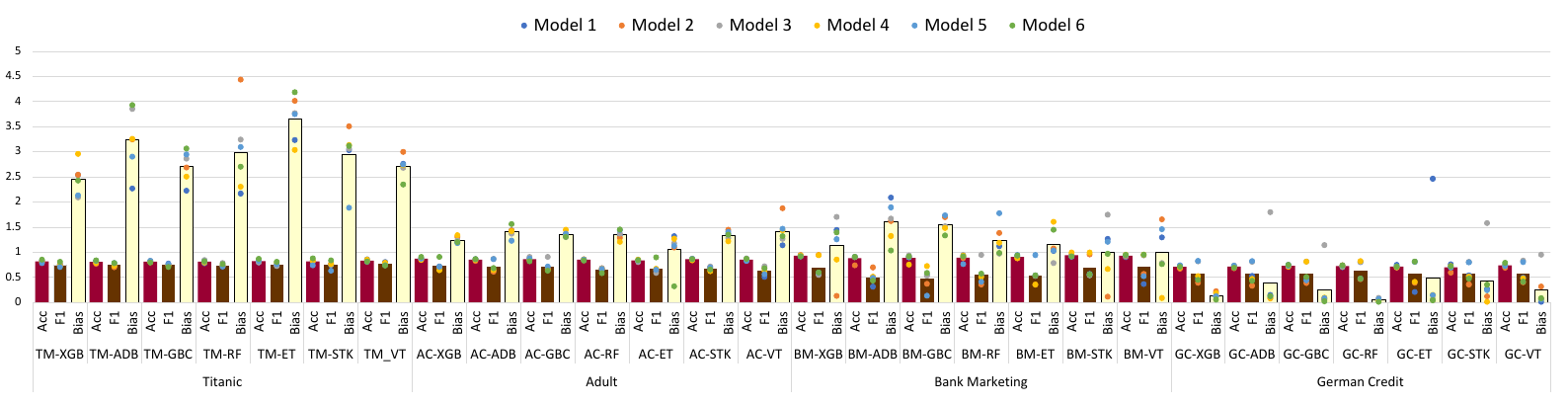}
    \caption{Cumulative bias and performance of all ensembles. The bars represent mean values, and the dots the models}
    \label{All_models}
\end{figure*}

Before understanding the composition of fairness in ensembles, we first investigate how different ensemble techniques impact fairness \textbf{(RQ1)}. Are certain ensemble classifiers more unfair? Does the architecture of an ensemble method (stacking, boosting, etc.) contribute to fairness? Does any particular ensemble classifier exhibit a better fairness-accuracy trade-off? To answer these questions, we experiment to evaluate the fairness of ensemble models using a diverse set of metrics. Table \ref{Table_Comparison_1} shows the mean fairness for all ensembles. Figure \ref{All_models} illustrates the cumulative fairness for all 168 models.

Our findings showed dataset-specific fairness patterns for ensemble models; however, some exhibited more unfairness than others. We used the Scott-Knott ranking test \cite{7194626} to compare the fairness and accuracy of the ensemble types and determine if the differences are significant. The test assigns a rank to the classifiers based on their performance, with a higher rank indicating better results. In our experiments, the classifiers were ranked from 1st to 5th (some with the same rank) for each metric.

\stepcounter{finding}
\begin{mdframed}[style = exampledefault]
\noindent \textbf{Finding \thefinding:} Among all the ensemble models, XGBoost exhibits the best accuracy-fairness trade-off.
\end{mdframed}
\vspace{-1em}

Table \ref{Table_Comparison_1} shows varying fairness performance among the ensemble classifiers across different datasets.  Interestingly, we observe that fairness can be highly inconsistent even within the same ensemble type. For instance, \textit{XGBoost} has the highest rank in 8 out of 10 fairness metrics for the highly biased \textit{Titanic} and \textit{Adult} datasets, with a mean fairness rank of 1.4 and 1.6, respectively. On the other hand, \textit{AdaBoost} has the lowest rank in 13 out of 16 fairness metrics across all the datasets. Additionally, we observe that \textit{XGBoost} stands out with high accuracy and fairness across all ensemble models, contrary to typical inverse behavior seen in ML models. 
For example, in the Titanic dataset, the average performance change for the XGBoost classifier in accuracy and f1 score is less than 0.01. However, their cumulative mean fairness is $14\%$ more than the next most fair model (\textit{GBC}) in Titanic. For the other datasets, we observe a similar pattern; however, the difference is lesser due to low unfairness in the dataset. Upon further investigation, we found that boosting method and base learner design is responsible for the fairer performance of \textit{XGBoost}. Homogeneous ensembles use decision trees as the base learner, and the construction of these trees differs among them. For example, the depth of the decision tree in \textit{AdaBoost}, \textit{GradientBoosting}, \textit{XGBoost} is one, three, and six, respectively. Lower tree depth means fewer features are selected, which has been shown to often increase unfairness \cite{9402057, 10.1145/3368089.3409704}. Importantly, we found that the fairness of an ensemble is determined by the composition of fairness within these base learners and the learning method (boosting, bagging, etc.). In the next section, we delve deeper into the properties of base learners to understand how to create fair ensembles.

\stepcounter{finding}
\begin{mdframed}[style = exampledefault]
\noindent \textbf{Finding \thefinding:} Fairness measures show more instability compared to accuracy metrics.
\end{mdframed}

\begin{figure}[t]
  \centering
    \includegraphics[width=\columnwidth]{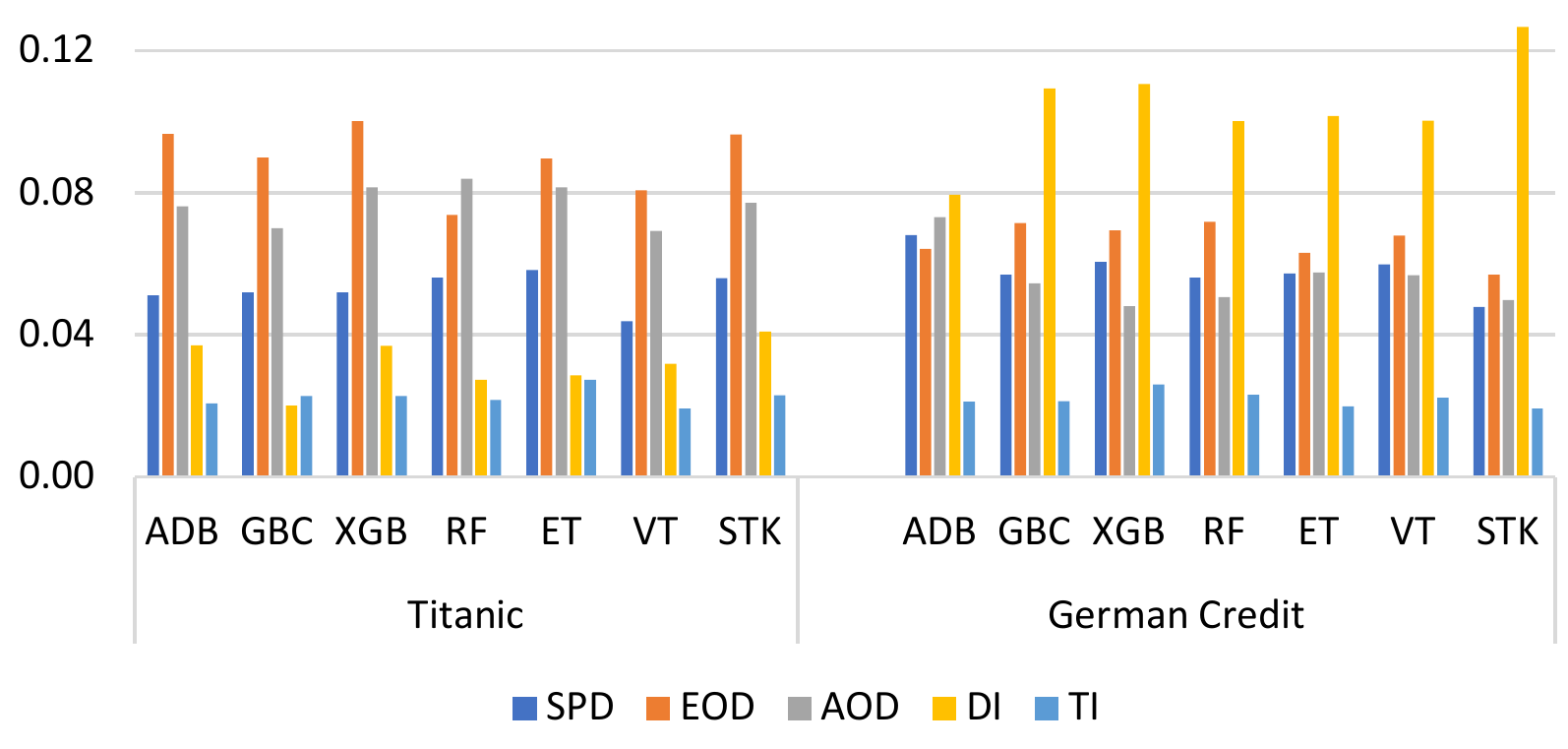}
    \caption{Standard deviation of fairness metrics over multiple experiments. Other datasets have very low standard deviation.}
    \label{fig:Std}
\end{figure}

\begin{figure*}[]
  \centering
    \includegraphics[width=\linewidth]{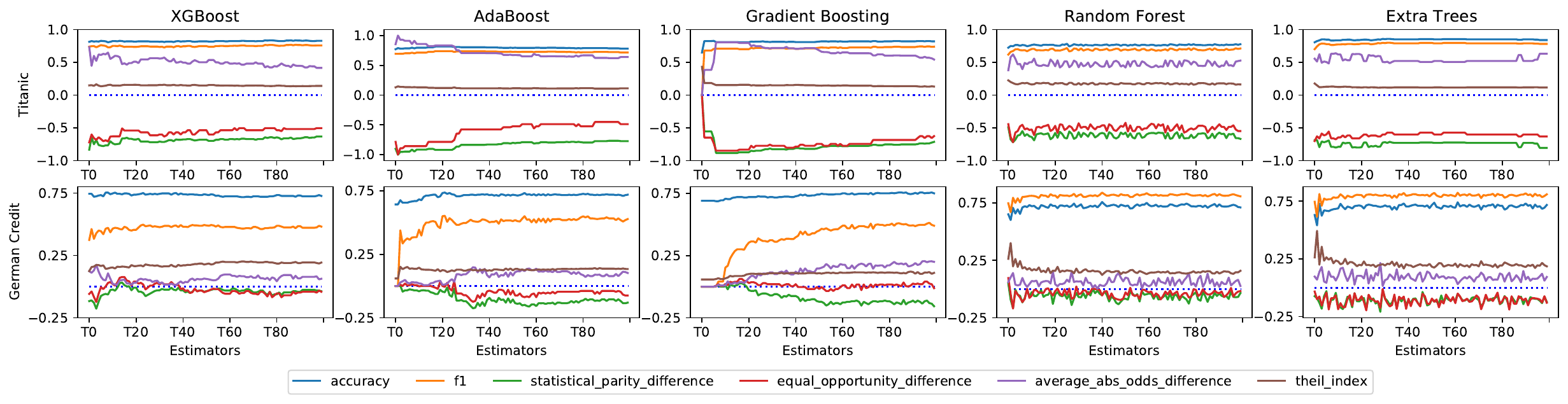}
    \caption{Composition of fairness in homogeneous ensembles with respect to base learners. Default no. of trees: T100}
    \label{Sequential_Ensembles}
\end{figure*}

Prior works have shown that ensembles improve the stability of accuracy metrics by aggregating multiple learners trained on subsets of data (bootstrap sampling) \cite{Bellamy2018AIF3, breiman1996bagging}. Intuitively, we explore the variance exhibited by fairness metrics in ensembles. Interestingly, despite aggregating multiple learners, the stability of fairness measures in ensembles still suffers, especially in smaller datasets. This is attributed to the change in data distribution after random train/test splits in smaller datasets \cite{10.1145/3287560.3287589, 10.1145/3368089.3409704}. For larger datasets (\textit{Adult, Bank Marketing}), the standard deviation for all fairness measures is less than 0.02. For smaller datasets, the average standard deviation of the metrics is shown in Figure \ref{fig:Std}. Firstly, we observe that the stability of fairness metrics remains consistent between all the ensembles for a specific dataset. Furthermore, we observed that group fairness measures exhibit higher variability than individual fairness measures (TI). Surprisingly, heterogeneous models also exhibit instability despite using dissimilar learners to reduce variance. From Figure \ref{Sequential_Ensembles}, we also see that the volatility in fairness metrics is greater than in accuracy metrics for homogeneous models. Given a random train/test split, it might cause the model to seem fairer than it is. Hence, even with improved stability in fairness compared to non-ensemble methods, developers should evaluate the training set and repeat training over multiple runs in ensembles.

\stepcounter{finding}
\begin{mdframed}[style = exampledefault]
\noindent \textbf{Finding \thefinding:} Libraries do not provide API support to measure fairness of base learners in ensembles
\end{mdframed}

Biswas and Rajan \cite{10.1145/3368089.3409704} discussed that hyperparameter optimization goals induce unfairness. In the case of heterogeneous ensemble models, the developer must carefully choose the number and type of individual base learners. Libraries do not provide any recommendations to developers, who try to select a diverse set of learners to improve accuracy. However, this might not always result in a fair ensemble. For instance, removing a GaussianNB learner from the \textit{BM-STK3} model improved its Statistical Parity Difference (SPD) from 0.13 to 0.11 while also increasing accuracy. Heterogeneous ensemble models, such as \textit{Voting} models that use weighted voting and \textit{Stacking} models that use a meta-learner to determine the best weighing configuration of learners, can be challenging to train fairly since libraries do not provide API support to measure the fairness of base learners, especially in combinations with other learners at the ensemble level. Hence, developers have little information on how to weigh and select individual learners, which can lead to unfair ensembles. Similarly, understanding fairness composition in base learners of homogeneous models can help the developer identify fairness issues such as bias in specific features (e.g., decision tree learners in random forest
randomly select features). Therefore, API support to measure fairness in base learners can help developers better understand \& detect unfairness in ensembles.

\subsection{How does fairness compose in ensembles?}

\label{5.3}

In this section, we investigate the composition of fairness in ensembles. We posit that the underlying unfairness of ensembles is a product of the composition of fairness in base learners and the learning method. All homogeneous models use a decision tree as the base learner, whereas heterogeneous models can be constructed with any ML classifier. We investigate how fairness composes in these base learners and how it is propagated by the learning techniques \textbf{(RQ2)}. 
In general, our findings show that the complexity of base learners significantly impacts the fairness of ensembles and that more research is needed to develop fair learning techniques in ensembles.

\stepcounter{finding}
\begin{mdframed}[style = exampledefault]
\noindent \textbf{Finding \thefinding:} 
The unfairness of homogeneous ensembles is caused by the complexity of the base learner and dataset characteristics.
\end{mdframed}

In Figure \ref{Sequential_Ensembles}, we plotted the most biased and the least biased homogeneous ensembles in our benchmark. We see how the fairness of the base learners directly contributes to the fairness of ensembles during training. However, we observe different fairness composition patterns in terms of the datasets. For example, in \textit{Titanic}, the fairness of the boosting ensembles improves, while the opposite is observed for some fairness metrics in \textit{German Credit}. The variation in fairness patterns is also seen in specific classifiers. From Figure \ref{All_models}, we observe that all \textit{TM-XGB} models show similar bias except \textit{TM-XGB4}. We investigate the difference in unfairness by comparing all the parameters of the base learner decision tree with the other \textit{XGB} models and found that \textit{TM-XGB4} uses a shallow decision tree with $max\_depth:2$, which is causing the unfairness to amplify. The model construct is shown below:

\vspace{1pt}

\lstset{style=mystyle}
\begin{lstlisting}[language=Python]
model = XGBClassifier(n_estimators= 500,max_depth=2, subsample=0.5, learning_rate=0.1)
\end{lstlisting}

\vspace{1pt}

\textit{MaxDepth} sets the maximum depth of the decision tree. The depth of the tree is defined as the number of splits (nodes), where the feature to be split is chosen based on the highest information gain among the features. Deeper trees are more complex and reduce errors \cite{10.1023/A:1022643204877}. For \textit{XGBoost} models, the default depth is 6. Our analysis showed that the protected attribute (\textit{Sex}) has the highest information gain among all the features in the \textit{Titanic} dataset. Therefore, the protected attribute is the most important feature to split on at the tree's root, resulting in a high degree of unfairness in \textit{TM-XGB4}.  
Base learners in all boosting models use the best feature to split, which improves accuracy. However, it has been shown that unfairness is encoded in specific features \cite{10.5555/3504035.3504042}. If these features are also among the best features of a dataset, a shallower ensemble will be more unfair due to a reduced number of features. This corroborates similar observations in the literature \cite{10.1145/3468264.3468536, 9402057}. We observe the same pattern for all boosting models. For example, \textit{AdaBoost} and \textit{GradientBoosting} exhibit more bias than \textit{XGBoost} because of shallower base learners (1 and 3, respectively). In Figure \ref{All_models}, \textit{TM-GBC1} is fairer because the learner is deeper (depth:5).

Finally, further analysis of the properties of a decision tree suggests that regularization parameters like min samples leaf and max-leaf nodes also impact tree depth, hence affecting the fairness of the ensemble. Therefore, it is important to carefully balance the complexity of the tree-based base learners for homogeneous models with the fairness outcomes, especially with the underlying properties of the data.

\begin{figure}[]
  \centering
    \includegraphics[width=\linewidth]{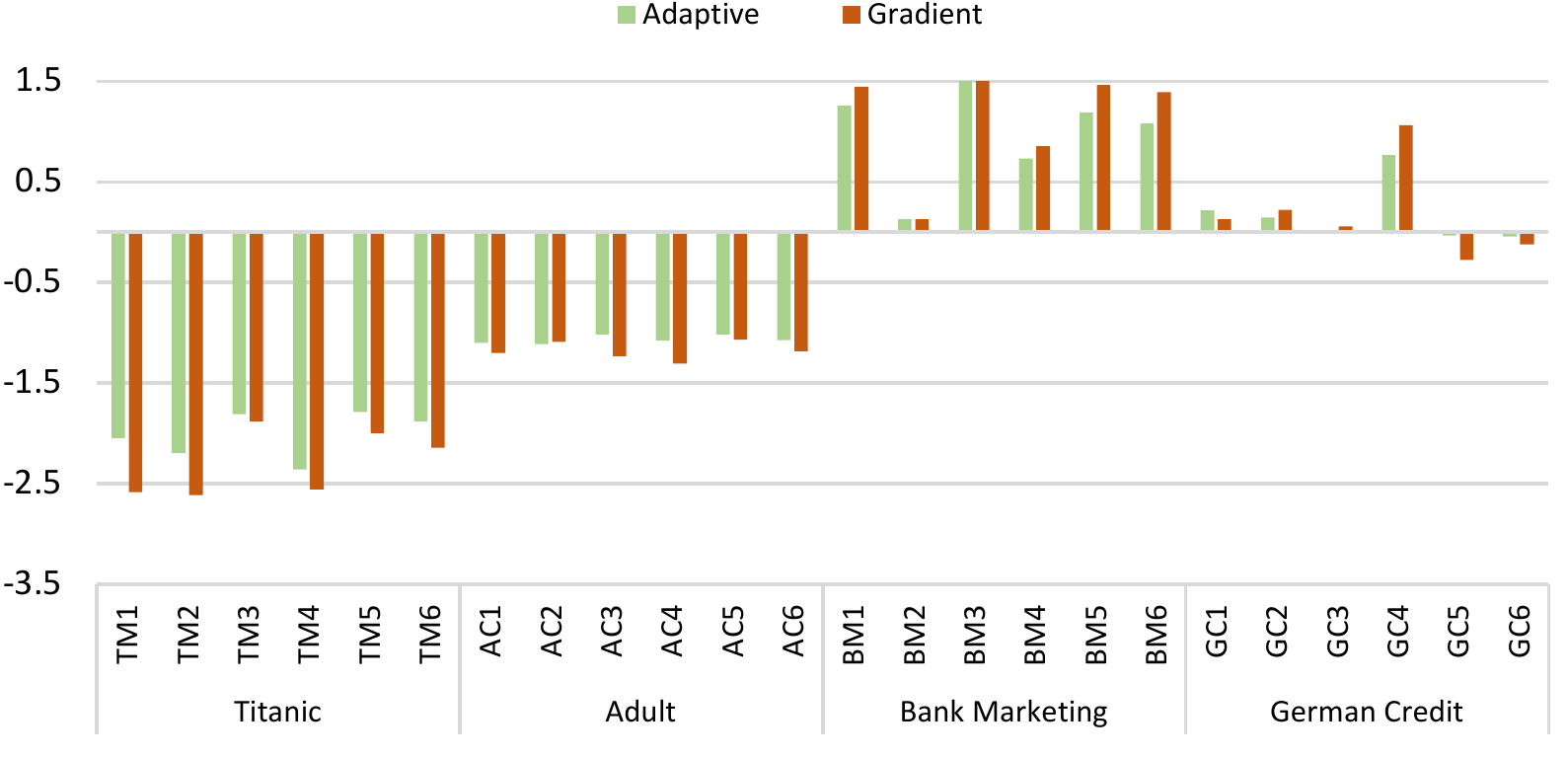}
    \caption{Cumulative fairness in adaptive vs. gradient learning}
    \label{GradientvsADB}
\end{figure}

\stepcounter{finding}
\begin{mdframed}[style = exampledefault]
\noindent \textbf{Finding \thefinding:} \textit{Gradient-based} composition propagates more unfairness compared to \textit{Adaptive} boosting models.

\end{mdframed}
We have established that base learners and the underlying data properties influence the unfairness of homogeneous models. However, boosting models also use a learning technique to improve the model's predictions sequentially. \textit{XGBoost} and \textit{GradientBoosting} models use gradient-based optimization and \textit{AdaBoost} uses an adaptive weighting technique. We compare these techniques by training the boosting models on the same base learner decision tree (depth:6). We only use \textit{XGBoost} and choose this depth in our experiment since our analysis (Table \ref{Table_Comparison_1}) showed that it is the most fair boosting model. The results are shown in Figure \ref{GradientvsADB}. For all the models except GC1, we see that adaptive learning is fairer than gradient optimization. We use the Scott-Knott rank test to test statistical significance. Accordingly, we observed that adaptive learning outperformed gradient optimization in all datasets except German Credit, where the difference was not statistically significant. Consequently, we can see that adaptive learning propagates less bias in highly biased datasets. Our analysis should help guide further research into designing fair learning techniques for boosting ensembles.

\section{Fair Ensembles Design}
\label{sec:fair-design}

\begin{table*}[!htbp]
  \centering
  \caption{Ensemble-related hyperparameters (HP) that can affect the design of fair ensembles}
  \setlength{\tabcolsep}{4pt} 
  \setlength\extrarowheight{1pt}
    \begin{tabular}{|l|l|c|c|c|c|c|c|c|c|}
    \hline
    \textbf{HP} & \textbf{Values} & \textbf{Default} & \textbf{ADB} & \textbf{GBC} & \textbf{XGB} & \textbf{RF} & \textbf{ET} & \textbf{Voting} & \textbf{Stacking} \\
    
    \hline
    n\_estimators & Total number of trees/boosting rounds & 100 & \textcolor[rgb]{ 0,  .69,  .314}{\checkmark} & \textcolor[rgb]{ 0,  .69,  .314}{\checkmark} & \textcolor[rgb]{ 0,  .69,  .314}{\checkmark} & \textcolor[rgb]{ 0,  .69,  .314}{\checkmark} & \textcolor[rgb]{ 0,  .69,  .314}{\checkmark} &  &  \\
    \hline
    booster  & Booster type: {gbtree, gblinear, dart}& gbtree &              & & \textcolor[rgb]{ 0,  .69,  .314}{\checkmark}     &       &       &       &  \\
    \hline
    bootstrap & Data sampling with replacement & RF: True, Others: False &       &   \textcolor[rgb]{ 0,  .69,  .314}{\checkmark}     &    \textcolor[rgb]{ 0,  .69,  .314}{\checkmark}    & \textcolor[rgb]{ 0,  .69,  .314}{\checkmark} & \textcolor[rgb]{ 0,  .69,  .314}{\checkmark} &       &  \\
    \hline
    voting & Voting Type: {soft, hard} & hard &       &       &             &       &       & \textcolor[rgb]{ 0,  .69,  .314}{\checkmark} &  \\
    \hline
    estimators & Base learners for the ensemble &  ADB: DecisionTree, Others: none & \textcolor[rgb]{ 0,  .69,  .314}{\checkmark}   &       &              &       &       & \textcolor[rgb]{ 0,  .69,  .314}{\checkmark} & \textcolor[rgb]{ 0,  .69,  .314}{\checkmark} \\
    \hline
    final\_estimator & Meta-learner to combine learner predictions & Logistic Regression &       &       &              &       &       &       & \textcolor[rgb]{ 0,  .69,  .314}{\checkmark} \\
    \hline
    \end{tabular}%
  \label{Parameters}%
\end{table*}%

In \S \ref{sec:fairness-in-ensembles}, we found that base learners of ensembles propagate bias. Many bias mitigation techniques applied during model training (\textit{inprocess}) have been successful \cite{10.1145/3368089.3409697,10.1145/3510003.3510202, 10.1145/3278721.3278779}. Techniques applied during the model training phase can assist developers in improving the fairness of ML software. These works have also established the role of hyperparameters in mitigating and amplifying fairness bugs (unfairness) in ML software. If we understand and identify what ensemble parameters and design choices affect the fairness, we can mitigate inherent bias in ensembles. Moreover, it will help developers, and libraries better explain fairness bugs in the ensemble hyperparameter space. This section explores the hyperparameter design space for ensembles to boost fairness performance. We have found that some hyperparameters directly affect the fairness of ensembles. Specifically, we evaluate how ensembles can be designed to be fair using ensemble hyperparameters summarized in Table \ref{Parameters}. We use the Scott-Knott test to determine the significance of our results. Our findings provide a comprehensive review of all ensemble hyperparameters.

\stepcounter{finding}
\begin{mdframed}[style = exampledefault]
\noindent \textbf{Finding \thefinding:} Developers should carefully choose dropout regularization to balance fairness and overfitting.
\end{mdframed}

\begin{figure}[]
  \centering
    \includegraphics[width=\columnwidth]{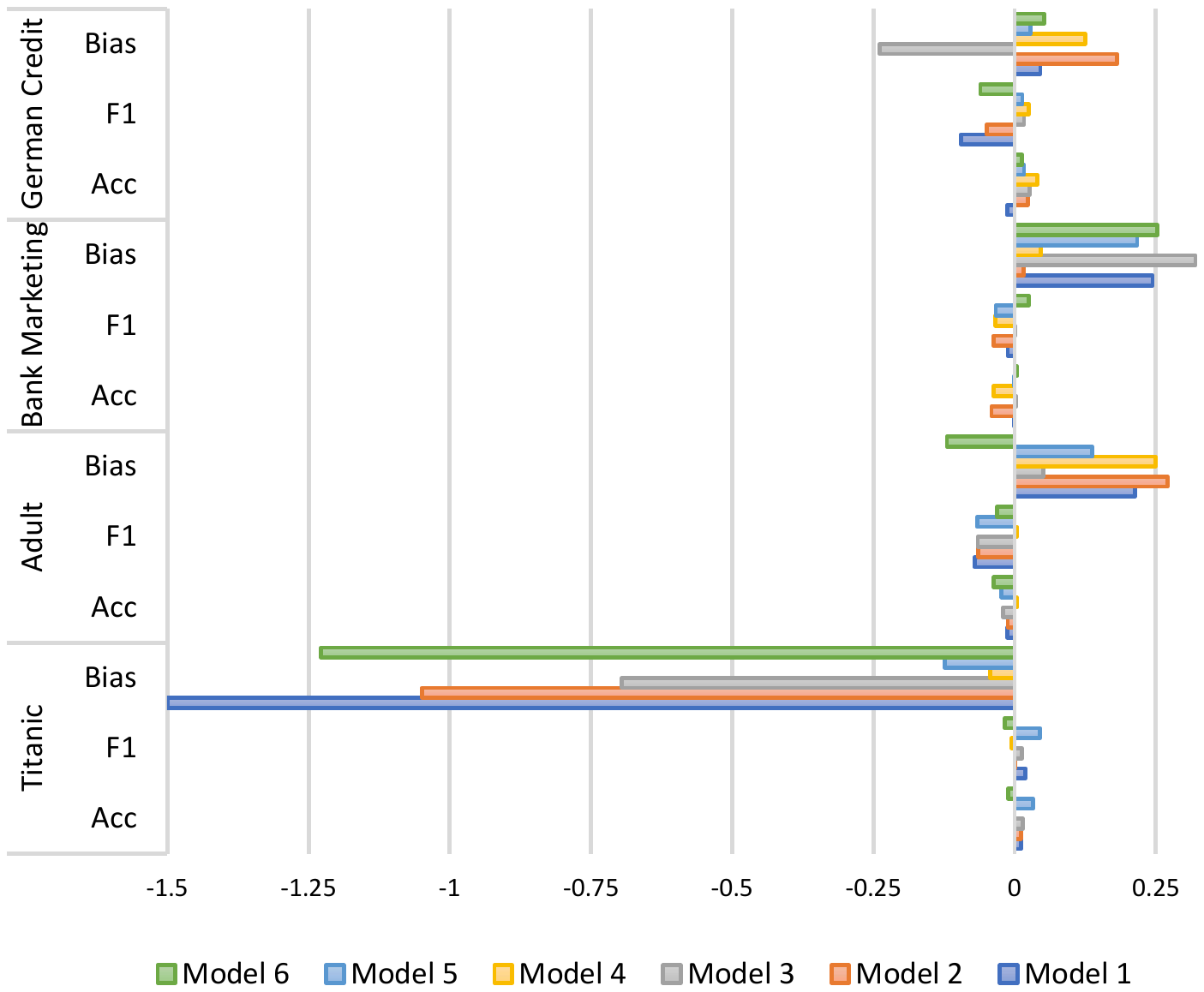}
    \caption{Change in accuracy and total bias when using DART. A negative value denotes lower fairness \& accuracy}
    \label{Dart:comparison}
\end{figure}

Our analysis shows that dropout impacts fairness in relation to the underlying data properties. Vinayak and Gilad-Bachrach \cite{rashmi2015dart} proposed DART, a dropout technique derived from deep neural networks, for boosted trees. An ensemble of boosted regression trees suffers from over-specification, i.e., the trees added at the end have little contribution to the final result \cite{rashmi2015dart}. \textit{DART} alleviates this by constructing the next tree from the residuals of a random sample of the previous trees. In \textit{XGBoost}, the \textit{rate-drop} ([0-1]) parameter controls this sampling rate. No trees are dropped on the lower end of this rate, while on the higher extreme, all trees are dropped. We investigate the efficacy of DART with \textit{ratedrop = 0.5}, in reducing unfairness in boosting models by comparing it with the default \textit{XGBoost} booster \textit{gbtree}. We analyze the change in performance and fairness of dropout in Figure \ref{Dart:comparison}.

From Figure \ref{Dart:comparison}, we can see that dropout can impact the fairness of boosting models. For example, in \textit{Adult} dataset, initial trees exhibit less unfairness than the latter. Using dropout, the subsequent trees only learn on a random sample of initial trees, which in this case are fairer. This improves the fairness of the models. The opposite is observed in \textit{Titanic} dataset. In both scenarios, the change in accuracy is less than 0.1, but a significant impact is seen on fairness. Therefore, developers should be cautious about the effect of regularization on fairness. More research is needed to understand the fairness-overfitting trade-off and develop fair regularization.

\stepcounter{finding}
\begin{mdframed}[style = exampledefault]
\noindent \textbf{Finding \thefinding:} Randomness in feature splitting does not improve fairness in bagging models.
\end{mdframed}

\textit{Random Forest} improves the variance of the model by introducing randomness to the process of model building by randomly selecting features. \textit{Extra Trees} introduces additional randomness by randomly finding the splits for each feature and then selecting the best split from them, i.e., independent of the target variable. In contrast, \textit{Random Forest} finds the best split for each feature which has been shown to improve accuracy \cite{geurts2006extremely}. However, no work has studied its effect on fairness. Here, we ask whether randomness at the feature splitting level causes bagging models to be unfair.

To investigate this, we compare \textit{Random Forest} and \textit{Extra Tree} models in our benchmark. We keep the rest of the parameters and data split the same. Each model is run ten times, and the mean is reported in Table \ref{tab:RFvsET}. For all datasets except \textit{German Credit}, the test showed that \textit{Extra Tree} models with random splits were more biased compared to optimal splits (\textit{RF}). This is a key finding because this suggests that a split chosen independently of the target is still more unfair than an optimal split. However, in the fairer dataset (\textit{German} Credit), we observe no difference in fairness. Regarding bias mitigation methods, our results suggest randomness in feature split-point might not be an effective way to tackle bias in decision tree-based models.

\begin{table}[t]
  \centering
  \caption{Mean total fairness in Random Forest (RF) and ExtraTrees (ET) models. $^*$ denotes the top rank based on the Scott-Knott significance rank test for \textit{each} dataset.}
  \resizebox{\columnwidth}{!}{
    \begin{tabular}{crrrrrrrr}
    \toprule
          & \multicolumn{2}{c}{\textbf{Titanic}} & \multicolumn{2}{c}{\textbf{Adult}} & \multicolumn{2}{c}{\textbf{Bank Marketing}} & \multicolumn{2}{c}{\textbf{German Credit}} \\
\cmidrule{1-9}          & RF$^{*}$    & ET    & RF$^{*}$    & ET    & RF$^{*}$    & ET    & RF    & ET \\
    \toprule
    Model1 & -2.04 & -1.96 & -1.33 & -1.32 & 1.16
 & 1.35  & 0.03  & 0.02 \\
    Model2 & -4.43 & -5.37 & -1.29 & -1.30 & 1.37
  & 1.41  & 0.43  & 0.29 \\
    Model3 & -3.38 & -4.13 & -1.32 & -1.45 & 0.96  & 1.07  & 0.02  & -0.04 \\
    Model4 & -2.42 & -2.46 & -1.21 & -1.29 & 1.36  & 1.35  & -0.16 & -0.09 \\
    Model5 & -2.98 & -3.38 & -1.48 & -1.51 & 1.61
  & 1.51
  & 0.00  & -0.04 \\
    Model6 & -2.61 & -2.61 & -1.47 & -1.41 & 0.98  & 1.26  & -0.04 & -0.28 \\
    \bottomrule
    \end{tabular}}%
  \label{tab:RFvsET}%
\end{table}%

\stepcounter{finding}
\begin{mdframed}[style = exampledefault]
\noindent \textbf{Finding \thefinding:} The uncertainty in
classifiers can have a large impact on fairness in voting classifiers. 
\end{mdframed}

A \textit{Voting} classifier is an ensemble method where the prediction is based on the probabilities of each base learner within the ensemble. Voting classifiers are of two types, \textit{Soft} and \textit{Hard} Voting. In hard Voting, the label with the majority of votes from the base learners is the final prediction, whereas, in soft voting, it is based on the average of the probabilities of each output class. If the average probability of a class is less than 0.5, class 0 is predicted, and vice-versa.
We investigate the effect of the voting type on fairness and found that the uncertainty in the model prediction can have a large impact on fairness. For instance, \textit{AC-VT5} uses soft voting with \textit{Logistic Regression (LR), Random Forest (RF), KNN}, and \textit{Decision Tree (DT)} as base classifiers. As shown in Table \ref{tab:voting}, \textit{DT} introduces significant unfairness when used in soft voting compared to hard. We found that \textit{DT} has an output class probability of either 1 or 0 while other classifiers are in the range [0,1]. 

\begin{figure}[h]
  \centering
    \includegraphics[width=\columnwidth]{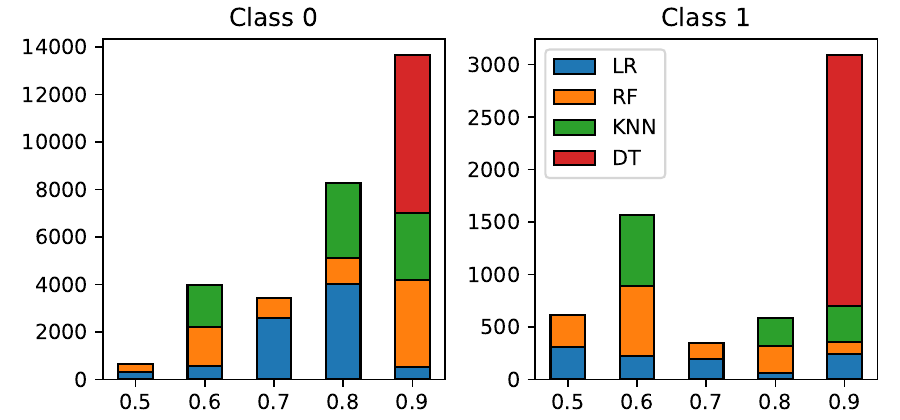}
    \caption{Frequency of output class probabilities for base learners in AC-VT5}
    \label{Voting}
\end{figure}

Figure \ref{Voting} shows the output probabilities for AC-VT5. We observe that DT has higher extreme probabilities compared to others. In this case, the average is skewed by the extreme probabilities of \textit{DT}. This changes the prediction for 558 out of 9049 test samples. And since \textit{DT} is less fair than other classifiers, the overall unfairness also increases. For hard voting, equal weight is given to each classifier. In that case, the other three classifiers, which are fairer, won the majority vote. For some models, soft voting was fairer than hard, e.g., \textit{AC-VT3}, which shows that base learners' uncertainties can impact fairness in both voting types. This suggests the need to develop frameworks to measure model uncertainties and their fairness at a component level to aid developers in designing fair voting ensembles. Our analysis should also encourage further research in fairness-aware weighting techniques to handle fairness issues arising from model uncertainties.

\begin{table}[]
  \centering
  \renewcommand{\tabcolsep}{1.3pt}
  \caption{Soft vs Hard Voting in \textit{AC-VT5}}
    \begin{tabular}{|c||r|r|r|r|r||r|r|r|r|r|}
    \hhline{-||----------}
    \multicolumn{1}{|c||}{} & \multicolumn{5}{c| |}{\cellcolor[HTML]{EFEFEF}\textbf{Soft}}     & \multicolumn{5}{c|}{\cellcolor[HTML]{EFEFEF}\textbf{Hard}} \\
    \hhline{-||-----||-----}
    \cellcolor[HTML]{EFEFEF} & \cellcolor[rgb]{ 1,  1,  1}\textbf{Voting} & \cellcolor[rgb]{ 1,  1,  1}\textbf{LR} & \cellcolor[rgb]{ 1,  1,  1}\textbf{RF} & \cellcolor[rgb]{ 1,  1,  1}\textbf{KNN} & \cellcolor[rgb]{ 1,  1,  1}\textbf{DT} & \cellcolor[rgb]{ 1,  1,  1}\textbf{Voting} & \cellcolor[rgb]{ 1,  1,  1}\textbf{LR} & \cellcolor[rgb]{ 1,  1,  1}\textbf{RF} & \cellcolor[rgb]{ 1,  1,  1}\textbf{KNN} & {\cellcolor[rgb]{ 1,  1,  1}\textbf{DT}}\\
    \hhline{-||-----||-----}
    \cellcolor[HTML]{EFEFEF}\textbf{Acc} & \cellcolor[rgb]{ 1,  1,  1}   0.83 & \cellcolor[rgb]{ 1,  1,  1}   0.78 & \cellcolor[rgb]{ 1,  1,  1}   0.81 & \cellcolor[rgb]{ 1,  1,  1}   0.81 & \cellcolor[rgb]{ 1,  1,  1}0.83 & \cellcolor[rgb]{ 1,  1,  1}0.81 & \cellcolor[rgb]{ 1,  1,  1}0.78 & \cellcolor[rgb]{ 1,  1,  1}0.81 & \cellcolor[rgb]{ 1,  1,  1}0.81 & \cellcolor[rgb]{ 1,  1,  1}0.81 \\
    \hhline{-||-----||-----}
    \cellcolor[HTML]{EFEFEF} \textbf{F1} & \cellcolor[rgb]{ 1,  1,  1}0.56 & \cellcolor[rgb]{ 1,  1,  1}0.41 & \cellcolor[rgb]{ 1,  1,  1}0.44 & \cellcolor[rgb]{ 1,  1,  1}0.43 & \cellcolor[rgb]{ 1,  1,  1}0.56 & \cellcolor[rgb]{ 1,  1,  1}0.44 & \cellcolor[rgb]{ 1,  1,  1}0.41 & \cellcolor[rgb]{ 1,  1,  1}0.41 & \cellcolor[rgb]{ 1,  1,  1}0.45 & \cellcolor[rgb]{ 1,  1,  1}0.44 \\
    \hhline{-||-----||-----}
    \cellcolor[HTML]{EFEFEF} \textbf{SPD} & \cellcolor[rgb]{ 1,  1,  1}-0.14 & \cellcolor[rgb]{ 1,  1,  1}-0.06 & \cellcolor[rgb]{ 1,  1,  1}-0.08 & \cellcolor[rgb]{ 1,  1,  1}-0.07 & \cellcolor[HTML]{FFCCC9}-0.14 & \cellcolor[rgb]{ 1,  1,  1}-0.07 & \cellcolor[rgb]{ 1,  1,  1}-0.06 & \cellcolor[rgb]{ 1,  1,  1}-0.06 & \cellcolor[rgb]{ 1,  1,  1}-0.08 & \cellcolor[rgb]{ .882,  1,  .906}-0.07 \\
    \hhline{-||-----||-----}
    \cellcolor[HTML]{EFEFEF} \textbf{EOD} & \cellcolor[rgb]{ 1,  1,  1}-0.14 & \cellcolor[rgb]{ 1,  1,  1}0 & \cellcolor[rgb]{ 1,  1,  1}0 & \cellcolor[rgb]{ 1,  1,  1}0 & \cellcolor[HTML]{FFCCC9}-0.14 & \cellcolor[rgb]{ 1,  1,  1}-0.03 & \cellcolor[rgb]{ 1,  1,  1}0 & \cellcolor[rgb]{ 1,  1,  1}-0.03 & \cellcolor[rgb]{ 1,  1,  1}-0.04 & \cellcolor[rgb]{ .882,  1,  .906}-0.03 \\
    \hhline{-||-----||-----}
    \cellcolor[HTML]{EFEFEF} \textbf{AOD} & \cellcolor[rgb]{ 1,  1,  1}0.1 & \cellcolor[rgb]{ 1,  1,  1}0.01 & \cellcolor[rgb]{ 1,  1,  1}0.02 & \cellcolor[rgb]{ 1,  1,  1}0.02 & \cellcolor[HTML]{FFCCC9}0.1 & \cellcolor[rgb]{ 1,  1,  1}0.02 & \cellcolor[rgb]{ 1,  1,  1}0.01 & \cellcolor[rgb]{ 1,  1,  1}0.02 & \cellcolor[rgb]{ 1,  1,  1}0.03 & \cellcolor[rgb]{ .882,  1,  .906}0.02 \\
    \hhline{-||-----||-----}
    \cellcolor[HTML]{EFEFEF} \textbf{DI} & \cellcolor[rgb]{ 1,  1,  1}-1.43 & \cellcolor[rgb]{ 1,  1,  1}-0.64 & \cellcolor[rgb]{ 1,  1,  1}-1.13 & \cellcolor[rgb]{ 1,  1,  1}-1.02 & \cellcolor[HTML]{FFCCC9}-1.43 & \cellcolor[rgb]{ 1,  1,  1}-1.08 & \cellcolor[rgb]{ 1,  1,  1}-0.64 & \cellcolor[rgb]{ 1,  1,  1}-1.11 & \cellcolor[rgb]{ 1,  1,  1}-1.02 & \cellcolor[rgb]{ .882,  1,  .906}-1.08 \\
    \hhline{-||-----||-----}
    \cellcolor[HTML]{EFEFEF} \textbf{TI} & \cellcolor[rgb]{ 1,  1,  1}0.17 & \cellcolor[rgb]{ 1,  1,  1}0.21 & \cellcolor[rgb]{ 1,  1,  1}0.2 & \cellcolor[rgb]{ 1,  1,  1}0.2 & \cellcolor[rgb]{ .882,  1,  .906}0.17 & \cellcolor[rgb]{ 1,  1,  1}0.2 & \cellcolor[rgb]{ 1,  1,  1}0.22 & \cellcolor[rgb]{ 1,  1,  1}0.21 & \cellcolor[rgb]{ 1,  1,  1}0.2 & \cellcolor[HTML]{FFCCC9}0.2 \\
    \hhline{-||-----||-----}
    \end{tabular}%
  \label{tab:voting}%
\end{table}%

\stepcounter{finding}
\begin{mdframed}[style = exampledefault]
\noindent \textbf{Finding \thefinding:} Two-layer stacking can significantly reduce unfairness.
\end{mdframed}



All of the \textit{Titanic} ML stacking models shown in Figure \ref{All_models} exhibit similar bias except \textit{TM-STK5}, which is the least biased model for all fairness metrics except Thiel Index (TI). On closer inspection, we found out that \textit{TM-STK5} uses a two-layered stacking approach where a second layer of base learners act as the meta-learner, which causes the model to be fairer. The model construct is shown below:

\lstset{style=mystyle, frame = single}
\begin{lstlisting}[language=Python,upquote=true]
layer_one_estimators = [('rf_1', RandomForestClassifier(n_estimators=40, random_state=42)),('knn_1', KNeighborsClassifier(n_neighbors=6))]
layer_two_estimators = [('rf_2', RandomForestClassifier(n_estimators=40, random_state=42)),('xg_2', XGBClassifier(objective = 'binary:logistic', colsample_bytree = 0.8, learning_rate = 0.3, max_depth = 7, min_child_weight = 3, n_estimators = 100, subsample = 0.6))]
layer_two = StackingClassifier(estimators=layer_two_estimators, final_estimator=XGBClassifier(n_estimators  = 100))
model =  StackingClassifier(estimators=layer_one_estimators,final_estimator=layer_two)}
\end{lstlisting}

\begin{figure}[]
  \centering
  \includegraphics[width=\columnwidth]{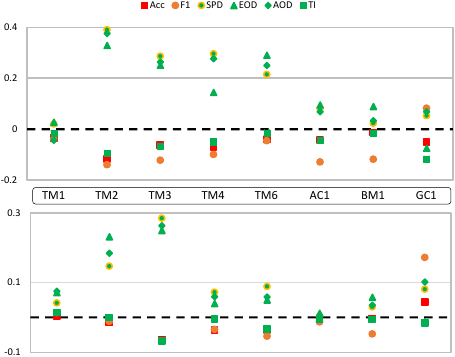}
    \caption{Performance \& fairness changes after using two-layered stacking (Top) and XGB as meta-learner (Bottom). A positive change indicates performance/fairness increase and vice-versa}
    \label{Fig:Stacking_Nested}
\end{figure}

We validate our finding by training all stacking models in our benchmark using this two-layered nested stacking approach. To ensure consistency, we did not change the kernel's feature set or any preprocessing method. The results are shown in Figure \ref{Fig:Stacking_Nested}. For all stacking models in our benchmark, every model significantly improved in all fairness measures except \textit{Thiel Index}, which is typical as previous works \cite{Chouldechova2017FairPW, https://doi.org/10.48550/arxiv.1703.09207} have shown that achieving fairness in terms of all fairness metrics is often difficult. Moreover, accuracy measures did not degrade significantly. For example, \textit{TM-STK6} improved \textit{SPD} scores by $28\%$ while accuracy dropped only $4.68\%$. Overall across all datasets, the accuracy score dropped by $6.50\%$ while the \textit{SPD} improved by $31.8\%$. 

In Stacking, the first layer uses a list of base learners to generate a set of predictions and a meta-learner to learn from them. However, instead of using a single classifier as the meta-learner, the predictions are fed into another set of base learners in a two-layered approach. This ensures that the outcome is not based on a single meta-learner. Therefore, the second layer of learners creates a new set of predictions, which are then fed into the second layer's final meta-classifier. We compare this approach to simply using an ensemble model (\textit{XGB}) as the meta-learner in default Stacking models and found similar patterns in fairness measures (Figure \ref{Fig:Stacking_Nested} bottom). Our results did not significantly vary between using other ensembles as the meta-learner. This supports observations that ensembles are fairer than standalone classifiers. Therefore, developers should use an ensemble as a meta-learner or the two-layer approach to improve the fairness of Stacking models.

\section{Discussion}
\label{sec:discussion}

In this study, we undertook the important task of understanding the composition of fairness in ensemble machine learning. Fairness of ML algorithms has been extensively studied, starting from empirical evaluation and identification \cite{10.1145/3368089.3409704,10.1145/3468264.3468536,9402057} to mitigation \cite{10.1145/3368089.3409697,10.1145/3468264.3468537,DBLP:journals/corr/abs-2111-02038,MAAT} and testing \cite{10.1145/3338906.3338937,10.1145/3106237.3106277,9793943,Fairtest}. However, no work has explored the composition of fairness in ensemble models, although ensembles cover the majority of prior SE works and open-source \cite{10.1145/3368089.3409704,10.1145/3468264.3468536,DBLP:journals/corr/abs-2006-08334}. We showed that considering ensemble models as monolithic classifiers leaves the opportunity to identify the root cause of unfairness. Consequently, our work has shown that fair ensembles can be designed without using bias mitigation techniques. Our research also identifies root causes of unfairness in different ensembles and their interplay with the input space in the pipeline \cite{10.1145/3510003.3510057}, which would guide fairness bug localization and repair in ensemble learning. For example, we report fairness patterns in individual learners that can induce bias in ensembles such as tree depth, minimum leaf node samples, etc. These can also be leveraged for fairness-improving interventions such as feature selection, data preprocessing, etc. Overall, our result would draw attention to the fairness of ensembles which are popular learning algorithms but mostly overlooked by the community.

Moreover, research in SE showed the impact of hyperparameters on fairness and their role in helping developers mitigate bias during model training \cite{10.1145/3510003.3510202, 10.1145/3468264.3468536, 10.1145/3368089.3409697}. We extend that to explore the hyperparameter space for ensembles to guide developers to design fair ensembles using currently available compositions and configurations. Our findings also made direct design suggestions for enhancing specific ensemble library APIs for feature splitting, dropout regularization, and fairness-accuracy trade-offs. This should encourage the development of fairness-aware regularization techniques and investigate the trade-off between fairness mitigation and overfitting. We found that many ensemble models do not have library support to monitor the fairness of individual learners. Finally, our work would encourage the development of tools and API support to improve the transparency of ML software to address fairness concerns.

\section{Threats to Validity}
\label{sec:threats}

\textbf{Benchmark:} We ensure the quality of the benchmark by collecting only high-quality kernels from Kaggle (at least five upvotes). Additionally, we only consider runnable models, include the protected attribute in training, and have an accuracy greater than 65\%, similar to \cite{10.1145/3368089.3409704}. Finally, we select the top 6 (upvotes) models for each ensemble type. 

\textbf{Sampling Bias:} To the best of our knowledge, this is the most extensive review of popular ensembles. Moreover, conclusions are supported by statistical tests across four datasets. However, they may change slightly if other datasets are used. 

\textbf{Generalizability:} To avoid the threat of non-generalized findings, we conduct experiments on four different datasets for each ensemble type and compare across multiple ensemble algorithms for both boosting and bagging. Moreover, we use multiple fairness metrics and verify our results by running the experiment multiple (ten) times and using the mean of the values.
\section{Related Work}
\label{sec:related}

\paragraph{Fairness in ML classification}The ML community has proposed multiple methods to measure \cite{10.1145/2090236.2090255,three-bayes, Chouldechova2017FairPW,10.1145/3278721.3278729, pmlr-v28-zemel13,NIPS2016_9d268236} and mitigate unfairness in ML models \cite{NIPS2016_dc4c44f6, NIPS2017_b8b9c74a, 10.1145/3278721.3278779,three-bayes, Chouldechova2017FairPW}. However, most of these works have focused on the theoretical evaluation of fairness. Recently, the SE community has increasingly shown interest in fairness in ML software \cite{fairnessinsoftware}. Empirical studies have investigated the characteristics of biased models and unfairness in ML pipelines, compared mitigation strategies and developer concerns about fairness \cite{10.1145/3368089.3409704,10.1145/3468264.3468536, 10.1145/3287560.3287589,10.1145/3290605.3300830}. Some research in SE has focused on fairness testing and verification and uncovering fairness violations \cite{9793943,10.1145/3106237.3106277, 10.1145/3338906.3338937, Fairtest, biswas23fairify}. Finally, a body of work has identified unfairness in data and proposed appropriate mitigation techniques \cite{10.1145/3468264.3468537, 10.1145/3368089.3409697, DBLP:journals/corr/abs-2111-02038, 9794106}. 


\paragraph{Ensemble Fairness}
Grgic-Hlaca \MakeLowercase{\textit{et al.}} \cite{Nina} investigated the impact of fairness in the random-selection-based ensemble. They showed theoretically that its fairness at the ensemble level is always fairer than its components. Wang \MakeLowercase{\textit{et al.}} \cite{49284} studied the composition of fairness in multi-component recommender systems and presented conditions under which individual components compose fairness. AdaFair \cite{8995403} proposed a fairness-aware AdaBoost model where unfairly classified instances were up-weighted. A recent work \cite{9462068} analyzed and compared seven ML models to show that ensembles were fairer than individual classifiers. Feffer \MakeLowercase{\textit{et al.}} \cite{IBMLale} conducted an empirical study to analyze modular ensembles. They developed a library to find the best configuration using any combination of ensembles and mitigators. In Fair Pipelines \cite{article}, the authors explored the propagation of fairness in multi-stage pipelines where a set of decisions impacts the final result, e.g., the hiring process. MAAT \cite{MAAT} proposes an ensemble approach to improve fairness performance by separately combining models optimized for fairness and accuracy. Finally, Tizpaz-Niari \MakeLowercase{\textit{et al.}} studied the parameter space of ML algorithms and its impact on fairness \cite{10.1145/3510003.3510202}. This work is the closest to our study; however, it proposed a testing approach to tune the parameters for achieving fairness and did not consider ensembles (except \textit{random forest}). Our work has focused on comprehensively evaluating fairness composition in all popular ensemble models and how the different algorithmic design configurations (parameters) impact fairness.

\section{Conclusion}
\label{sec:conclusion}

Ensembles are widely used for predictive tasks due to superior performance. However, most approaches to measuring fairness and mitigation focus on single classifiers. In this paper, we conduct an empirical study to evaluate the composition of fairness in popular ensemble techniques. The results showed that base learners induce bias in ensembles and that we can mitigate inherent bias in ensembles by using certain base learner configurations and appropriate parameters. Lastly, works have shown the need to support developers during model training in mitigating bias. Our analysis of the hyperparameter space should help developers build fairness-aware ensembles and automated tools to detect bias in ensembles.

\section*{Acknowledgment}
This work was supported in part by US NSF grants CCF-19-34884, CCF-22-23812, and CNS-21-20448. We also thank the anonymous reviewers for their insightful comments. All opinions are of the authors and do not reflect the view of sponsors.

\renewcommand{\bibfont}{\footnotesize}
\bibliographystyle{IEEEtran}
\bibliography{sample-base}


\end{document}